\documentclass[a4paper,fleqn]{cas-dc}

\usepackage[numbers]{natbib}
\usepackage{algorithmic}
\usepackage{algorithm}
\usepackage{bbding}
\usepackage{color}
\usepackage{xcolor}
\def\tsc#1{\csdef{#1}{\textsc{\lowercase{#1}}\xspace}}
\tsc{WGM}
\tsc{QE}

\begin{document}
\let\WriteBookmarks\relax
\def\floatpagepagefraction{1}
\def\textpagefraction{.001}
\let\printorcid\relax

\shortauthors{Hao Dong et~al.}  

\title [mode = title]{PGSO: Prompt-based Generative Sequence Optimization Network for Aspect-based Sentiment Analysis}  



%

\author[1]{\textcolor{black}{Hao Dong}}


\fnmark[1]

\ead{m202177083@hust.edu.cn}



\affiliation[1]{organization={School of Computer Science and Technology},
            addressline={Huazhong University of Science and Technology (HUST)}, 
            city={Wuhan},
            postcode={430074}, 
            state={Hubei},
            country={China}}

\author[2]{\textcolor{black}{Wei Wei}}


\ead{weiw@hust.edu.cn}


\cormark[1]
\affiliation[2]{organization={School of Computer Science and Technology},
            addressline={Huazhong University of Science and Technology (HUST)}, 
            city={Wuhan},
            postcode={430074}, 
            state={Hubei},
            country={China}}

\cortext[1]{Corresponding author}



\begin{abstract}
Recently, generative pre-training based models have demonstrated remarkable results on Aspect-based Sentiment Analysis (ABSA) task. However, previous works overemphasize crafting various templates to paraphrase training targets for enhanced decoding, ignoring the internal optimizations on generative models. Despite notable results achieved by these target-oriented optimization methods, they struggle with the complicated long texts since the implicit long-distance relation, e.g., aspect-opinion relation, is difficult to extract under the position embedding mechanism in generative models. Thus, in this paper, we first clarify the causes of the problem and introduce two sequence optimization strategies: the rule-based static optimization and the score-based dynamic optimization. The rule-based approach relies on handcraft priority of dependency relation to reorder the context, while the score-based algorithm dynamically regulates the contextual sequence by calculating word position scores using neural network. Based on the dynamic optimization structure, we further propose a unified Prompt-based Generative Sequence Optimization network (named PGSO), which jointly optimizes the training target as well as the generative model. Specifically, PGSO contains two components, namely, \emph{prompt construction} and \emph{sequence regulator}. The former constructs a task-specific prompt based on unsupervised training objects to fully utilize the pre-trained model. The latter jointly leverages semantic, syntactic and original-sequence information to dynamically regulate contextual sequence. Our experiments conducted on four ABSA tasks across multiple benchmarks indicate that PGSO outperforms state-of-the-art methods, with an average improvement of 3.52\% in F1 score.
\end{abstract}



\maketitle

\section{Introduction}\label{Introduce}
Aspect-based sentiment analysis (ABSA) focuses on mining detailed sentiment information related to specific aspects. There are four fundamental sentiment elements: aspect category (\emph{c}), aspect term (\emph{a}), opinion term (\emph{o}), and sentiment polarity (\emph{s}) \cite{7286808,10.1145/3616855.3635775,chauhan2023aspect} . Taking the sentence ``\emph{The pizza is tasty.}'' as an example, the corresponding elements are ``\emph{pizza}'', ``\emph{food quality}'', ``\emph{tasty}'' and ``\emph{positive}'', respectively. 
As shown in Table \ref{tab:ABSA output}, ABSA can be categorized into multiple tasks depending on the combination of various elements to be extracted.

 In general, ABSA tasks are formulated as discriminative manners by designing task-specific classification networks \cite{li-etal-2019-exploiting,huang2024flexibly,wu2023improving} or labeling strategies \cite{shi2023syntax,10394369}. However, these methods suffer from poor transfer-ability between different ABSA tasks due to these well-designed classifiers or strategies. Therefore, unified generative pre-trained based methods, especially on T5, gradually become the theme of ABSA tasks. GAS-T5 \cite{zhang-etal-2021-towards-generative} adopts T5 as the backbone model to tackle ABSA tasks with two styles of transferring paradigms. Zhang et al. \cite{zhang-etal-2021-aspect-sentiment} introduce a paradigm to conceptualize the quadruplet extraction (ASQP) as a paraphrase generation problem. Similarly, Mao et al. \cite{mao-etal-2022-seq2path} treat multiple sentiment tuples as a path of the tree, predicting targets independently. Despite remarkable results achieved by these target-oriented optimization methods, they still suffer from the inability of handling the complicated long texts, since they ignore the insufficiency of the generative model itself. 
\begin{table}[t]
 \caption{Output of different ABSA tasks}
    \centering {    
    \begin{tabular}{l c c}\hline
    \multicolumn{1}{c}{Task}  & \multicolumn{1}{c}{Abbr}&\multicolumn{1}{c}{Output} \\\hline
    \makecell{Unified Aspect-Based \\Sentiment Analysis}&UABSA & (\emph{a}\emph{,}\emph{ s}) \\
    \makecell{Aspect Sentiment \\Triplet Extraction} & ASTE  & (\emph{a}\emph{,}\emph{ o}\emph{,}\emph{ s}) \\
    \makecell{Target Aspect \\Sentiment Detection}& TASD   & (\emph{c}\emph{,}\emph{ a}\emph{,}\emph{ s})\\
    \makecell{Aspect Category\\Opinion Sentiment}&ACOS & (\emph{c}\emph{,}\emph{ a}\emph{,}\emph{ o}\emph{,}\emph{ s})\\
    \hline
    \end{tabular}
    }
    
    \label{tab:ABSA output}
\end{table}
\begin{table}[t]
\centering\caption{Error analysis of BERT-based \textbf{BMRC} model and T5-based \textbf{GAS-T5} model. For each sentence, the aspects and opinions are displayed in \textbf{bold}.}\label{Table:Error analysis}
	\begin{tabular}{c | c c}
            \hline
		Sentences & BMRC & GAS-T5 \\\hline
  \makecell{The \textbf{pizza} itself is not exactly\\ the best I 've had EVER, \\but still pretty \textbf{good} .}&  \makecell{Wrong\\prediction} &\makecell{Wrong\\prediction} \\\hline
  \makecell{The \textbf{scallops} are apparently \\cooked in a black olive butter\\ which really makes them \textbf{unique}\\(not to mention \textbf{tasty} ) .}&\makecell{Wrong\\prediction} &\makecell{Redundant\\prediction}\\\hline
	\end{tabular}
\end{table}

To provide further clarity on the issue discussed above, we present an error analysis comparing a BERT-based discriminative BMRC model \cite{Chen_Wang_Liu_Wang_2021} and a T5-based generative GAS-T5 model \cite{zhang-etal-2021-towards-generative} in Table \ref{Table:Error analysis}. Taking the first sentence as example, the aspect term is ``\emph{pizza}'' to the corresponding opinion term ``\emph{good}''. Nevertheless, extracting this long-distance aspect-opinion relation is challenging for both discriminative and generative models as they make incorrect prediction. As for the second sentence, the discriminative approach still struggles with extracting the correct relationships. While GAS-T5 successfully identifies the connection between ``\emph{scallops}'' and ``\emph{tasty}'', it unexpected pairs ``\emph{black olive butter}'' and ``\emph{tasty}'' as well.
 
To alleviate the issue, inspired by previous discriminative models, some works propose introducing syntax information to boost long-distance relation extraction\cite{phan-ogunbona-2020-modelling,10448322}. However, as compared to the various well-designed discriminative classifiers, a common generative-based decoder is usually pre-trained with fixed structure. Yu et al. \cite{yu2023syngen} advocate for the integration of syntax to optimize the contextual representations for better generation. Nevertheless, this paradigm leads to the sub-optimal performance owing to the well-known semantic gap and the potential noisy propagation. Consequently, how to effectively exploit the syntax for enhanced modeling long-distance relations in generative models is still an open problem.

The aforementioned problem motivates us to investigate a brand-new syntax-based approach for enhancing long-distance relation extractions, which is different from conventional target-oriented optimization methods. Given that relative position embedding exclusively pertains to the distance between the key and query (detailed theory will be illustrated in Section \ref{Preliminary}), a potential strategy to address the problem is regulating the contextual sequence to reduce the distance of concerned association (aspect-opinion relation). Meanwhile, the preservation of contextual representations contributes to mitigate the semantic gap from interference to self-attention calculations.

Therefore, based on the above viewpoint, we propose two contextual sequence optimization methods, named rule-based static optimization and score-based dynamic optimization respectively. The former regulates the contextual sequence based on the pre-defined rule, while the latter introduces the novel score-based structure to leverage the syntax information to dynamically regulate the context sequence. Based on the dynamic approach, we further propose an end-to-end Prompt-based Generative Sequence Optimization Network (named PGSO), which jointly optimizes the training target as well as the generative model. The proposed framework comprises two integral components: 1) \textbf{Prompt Construction} scheme, severing as target-oriented optimization method, which transforms the textual sequence generation task into cloze task to fully utilize the proposed model. For efficiency, we exclusively employ a straightforward fixed-template semantic prompt and a one-shot prompt. 2) \textbf{Sequence Regulator} module, operating as model-oriented optimization method, which dynamically rearranges context to boost extracting long-distance relations, especially aspect-opinion relations. Specifically, we design a score-based re-ranking scheme, transforming the original sequence optimization problem into a more easily modeled score permutation problem, thereby achieving dynamic regulation of the model's contextual sequence and effectively enhancing the model's ability in modeling long-distance dependency relations.

In summary, our contributions are as follows:
\begin{enumerate}
    \item To the best of our knowledge, this is the first work to raise the inability of long-distance relation extraction for generative models in ABSA tasks both conceptually and empirically. Furthermore, we propose two innovative contextual sequence optimization strategies, named rule-based static method and score-based dynamic method, to address the mentioned limitation.
    \item We propose a novel end-to-end score-based generative sequence optimization model, PGSO, which jointly optimizes task targets and pre-trained language model (PLM). In detail, the introduction of prompts transform the original generation task into a cloze-style, which aligns more closely with the tasks encountered during pre-training task. Meanwhile, by integrating new sequence regulator module, we dynamically optimize contextual sequence, thereby enhancing performance in long-distance relation extraction.
    \item Extensive experiments conducted on four ABSA tasks over 12 datasets demonstrate that our proposed model achieves state-of-the-art performance. To ensure a comprehensive evaluation for the proposed model, we also conduct an ablation study along with error and complexity analysis.
\end{enumerate}

\textbf{Roadmap.} The remaining of the paper is organized as follows.
In Section \ref{Related work}, we conduct a comprehensive review of the existing works. Section \ref{Preliminary} provides an overview of ABSA tasks, highlighting the inadequacies of the self-attention mechanism in the T5 model. Additionally, we delve into an analysis of the challenges posed by common syntax-based methods. The two sequence optimization methods and the architecture of our proposed network are introduced in Section \ref{Methodology}. Section \ref{Experiment} presents the quantitative results on benchmarks, and Section \ref{Analysis} includes an analysis and case sharing. Finally, Section \ref{Conclusion} serves as the conclusion, summarizing the key findings of the paper.

\section{Related Work}\label{Related work}
In this section, we make a comprehensive review of previous works, and point out the current problems of target-oriented optimization and model-oriented optimization.

Early studies for the ABSA tasks are concentrated on single sentiment element extractions such as Aspect Term Extraction (ATE) task \cite{10.1145/1014052.1014073}, or predicting the sentiment polarity of aspect term \cite{li-etal-2021-dual-graph}. Lately, some researchers propose multiple sentiment elements extractions, like pair, triplet even quadruplet. For pair extrication, the Unified Aspect-Based Sentiment Analysis (UABSA) task \cite{luo-etal-2019-doer} tries to jointly extracts the aspect term and predict its corresponding sentiment polarities. For triplet extraction, the primary tasks are focused on Target Aspect Sentiment Detection (TASD) \cite{DBLP:conf/aaai/WanYDLQP20} and Aspect Sentiment Triplet Extraction (ASTE) \cite{stanovich2000individual}. For quadruplet extraction, Aspect Category Opinion Sentiment (ACOS) \cite{cai-etal-2021-aspect} requires to predict four mentioned sentiment elements simultaneously.
\begin{table*}[pos=!ht]
\centering\caption{Training targets of existing works. They conventionally formulate ASBA tasks as standard text generations, employing static delimiters like commas or brackets.Taking ``\emph{Decent wine at reasonable prices.}'' and ASTE task as the example.}\label{tab:Training Targets}
\scalebox{0.95}{
\begin{tabular}{c | c } \hline
\multicolumn{1}{c|}{\textbf{Model}} & \multicolumn{1}{c}{\textbf{Training Targets}} \\\hline
GAS-T5 (Extraction Style) \cite{zhang-etal-2021-towards-generative} & (wine, positive, decent); (prices, positive, reasonable)\\
ParaPhrase \cite{zhang-etal-2021-aspect-sentiment}	& Food quality is great because wine is decent [SSEP] Price is great because prices is reasonable. \\
DLO \cite{hu-etal-2022-improving-aspect} & (decent, wine, positive); (reasonable, prices, positive) \\
Seq2Path \cite{mao-etal-2022-seq2path}& wine | decent | positive |||| prices | reasonable | positive\\
MvP ($Seq_i$) \cite{gou-etal-2023-mvp}& [S] positive [A] wine [O] decent [SSEP] [S] positive [A] prices [S] reasonable.\\\hline
\end{tabular}
}
\end{table*}

Since the sentiment polarity in Aspect-based Sentiment Analysis task belongs to three-element set (i.e., positive, negative, neutral), ABSA tasks are usually formulated as discriminative manners. Zhang et al. \cite{zhang-etal-2019-aspect} constructs a GCN over dependency tree to exploit syntactic information, boosting ABSA performance. Tang et al. \cite{tang-etal-2020-dependency} propose dual-transformer structure to jointly consider semantic and syntacitc channel. Liang et al. \cite{liang-etal-2022-bisyn} propose a syntax-aware framework to fully leverage syntax information of constituent tree based on BERT model. Gu et al. \cite{gu2023integrating} propose a graph convolutional network that fuses external sentiment knowledge to improve the ABSA performance. Tiwari et al. \cite{10084294} propose an adversarial anylysis baed on BERT model. Yadav et al. \cite{yadav2021positionless} simplify positional embedding calculation process with Bi-GRU structure. Zhang et al. \cite{zhang2022complete} propose a two-stage framework to solve compound ABSA tasks. Chen et al. \cite{chen-etal-2022-enhanced} propose an enhanced multi-channel GCN network for ASTE task. Despite significant results achieved by these discriminative manners, they exhibit poor transfer-ability across various sub-tasks due to their classifiers are designed for certain specific sentiment elements or ABSA sub-tasks.

Recently, end-to-end generative based approaches have been widely used to tackle various ABSA tasks uniformly. Different from discriminative manners, generation-based methods are not confined to specific tasks, predicting all the sentiment elements in an auto-regressive style. Meanwhile, generative models consider the rich label semantics, and do not require an extra task-specific classifier. Zhang et al. \cite{zhang-etal-2021-towards-generative} propose GAS-T5 framework, which is the first work to adopt T5 as backbone model to tackle ABSA tasks with two paradigms, namely annotation and extraction style, formulating each ABSA task as text generation task. Based on this research, numerous optimization methods based on the generative model are emerged to improve the performance. Specifically, optimizations on generative models are mainly categorized into two directions: target-oriented optimization method and model-oriented method.

\subsection{Target-oriented optimization}\label{too}

Target-oriented optimization methods require designing various templates to paraphrase training targets in terms of objective order or format. Zhang et al. \cite{zhang-etal-2021-aspect-sentiment} propose a method that transfers the quadruplet or triplet extraction into paraphrase generation with pre-defined templates, explicitly modeling the semantic relation between the sentiment element. Hu et al. \cite{hu-etal-2022-improving-aspect} investigate the order of generated sentiment elements, and try to find the best sequences for each task. Gao et al. \cite{gao-etal-2022-lego} combine sentiment element prompts to tackle various ABSA tasks. Mao et al. \cite{mao-etal-2022-seq2path} separate training targets independently, treating sentiment tuple as a path of a tree, and select valid paths via discriminative word with beam search technology. Gou et al. \cite{gou-etal-2023-mvp} jointly consider different orders of targets, solving ABSA tasks from different perspectives. 

As shown in Table \ref{tab:Training Targets}, most existing works conventionally formulate ABSA tasks as standard text generation, employing static delimiters like commas or brackets. However, these methods usually have significant differences between pre-training tasks and the ABSA training tasks, leading to performance decline when transferred to ABSA tasks. Meanwhile, these methods concentrate on designing target-oriented templates, paraphrasing the sentiment elements from various perspectives to enable a more comprehensive understanding of ABSA tasks by pre-trained language models, thereby improving the quality of generation. Despite their effectiveness, they heavily rely on the inherent structure and generation capability of the original pre-trained language model, ignoring the inability in capturing long-distance aspect-opinion relations due to the insufficiency of the generative model itself.

\subsection{Model-oriented optimization}
Compared with various target-oriented optimization methods, model-oriented optimization methods are relatively scarce. Yu et al. \cite{yu2023syngen} design a dual-channel encoder and a pointer decoder based on the BART \cite{lewis2019bart} (adopting similar structure with T5), aiming to improve the alignment between aspects and opinions. Fei et al. \cite{fei2022lasuie} investigate a structure-aware generative language model that leverages syntactic representations for better unified information extraction including ABSA tasks. Different from previous works, our model introduces a novel plug-in sequence regulator located between the encoder and decoder. In tandem with the architectural enhancement, we also overwrite the model's default generative function.

\section{Preliminary}\label{Preliminary}
In this section, we make a detailed explanation in theory (section 3.2) and experimentally (section 3.3) to better understand the mentioned issue.

\subsection{Problem Definition}
 ABSA task aims to identify and analyze sentiment associated with specific aspects within the given texts. Given an input sentence $\emph{s}=\left\{w_i\right\}_{n}$, where $n$ is the length of text. ABSA task is to predict the sentiment tuples set $T=\left\{t_i\right\}_{m}$, where $m$ stands for the quantity of the sentiment tuples contained in the input text. Based on different task requirements in Table \ref{tab:ABSA output}, each tuple $t_i$ is consisted of several sentiment fundamental elements. Taking ACOS task as an example, the sentiment tuple $t_i=(c_i,a_i,o_i,s_i)$, where $c_i$, $a_i$, $o_i$, $s_i$ represent aspect category, aspect term, opinion term and sentiment polarity consisted in the $i$-th tuple respectively.
 \begin{figure}[ht]
\centering	\includegraphics[width=0.49\textwidth,height=0.70\linewidth]{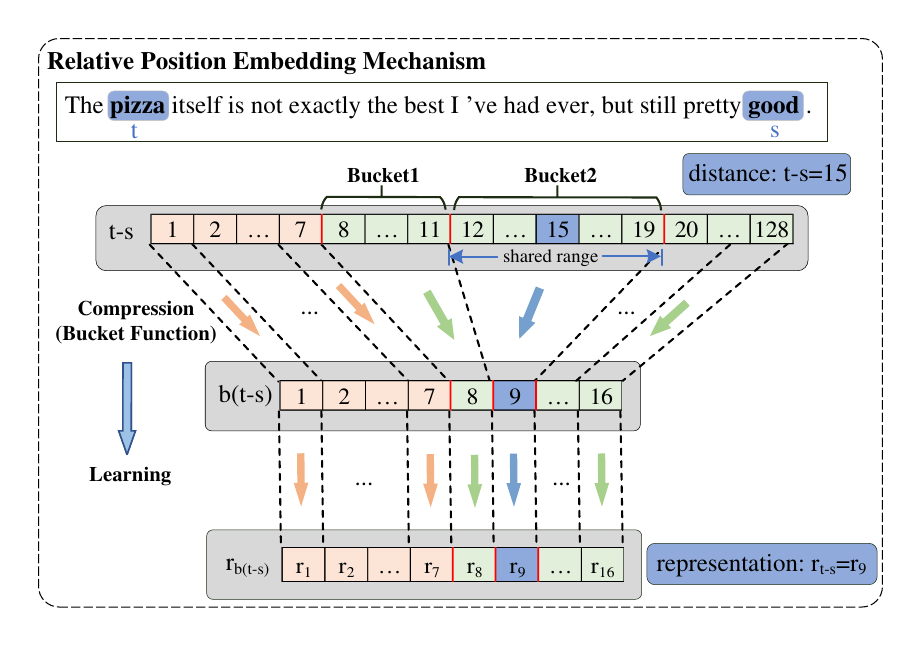}
	\caption{Part of the relative position embedding mechanism. The initial distance between the words \emph{pizza} and \emph{good} is measured at 15. This distance is subsequently reduced to 9 through a compression process. As a result, the precise positional information provided by the input text is diminished. Notably, with larger distances, the \emph{bucket} or range is also becomes wider.}
	\label{Fig:position function}
\end{figure}
\subsection{Relative position mechanism}
T5 \cite{10.5555/3455716.3455856} is a typical generative pre-trained model, employing Transformer-based encoder-decoder structure. Since self-attention is order-independent, an explicit position signal are provided to the calculation process. For efficiency, T5 adopts the simplified form of relative position embedding, which can be formulated as follows, 
\begin{equation}
\boldsymbol{A}_{{t}, {s}}=(\boldsymbol{W}_Q \boldsymbol{x}_{t} )^T\boldsymbol{W}_K \boldsymbol{x}_{s}+r_{b({t}-{s})}
\label{attention cal}
\end{equation}
where the $\boldsymbol{x}_{t}$ and $\boldsymbol{x}_{s}$ indicates the representations of the query and key words, and $t-s$ represents the relative distance between the key and query words. $b(.)$ represents the bucket function. In the Equation \ref{attention cal}, the former represents the common self-attention calculation, while the latter signifies the incorporation of position signals. As illustrated in Figure \ref{Fig:position function}, the relative position embedding mechanism comprises two distinct processes: compression and learning. In the compression process, a bucket function is employed to generate a fixed number of offsets. Specifically, short key-query offsets are retained, whereas long offsets are truncated using a pre-defined list, whose shared range expands with the increasing offset values. In the learning process, a shared scale $r_{b({t}-{s})}$ is acquired for each offset, contributing to the computation of attention weights. Still taking ``\emph{The pizza itself is not exactly the best I 've had ever, but still pretty good.}'' as an example, the original distance (15) between ``\emph{pizza}'' and ``\emph{good}'' will be clipped to 9 with neighbors by bucket function, which results in a loss of precise position information between aspect and opinion word. Hence, T5-based target-oriented optimization methods mentioned in the Section \ref{too} encounter challenges in addressing long-distance aspect-opinion relations. 

\begin{table}[ht]
\centering
\caption{Comparison between the contextual representation optimization method and the baselines in ASTE task. "CRO" represents the \textbf{C}ontextual \textbf{R}epresentation \textbf{O}ptimization method. The best performances are in \textbf{bold}, and second-best are \underline{underlined}.}\label{tab:Reprensatation optimization}
\begin{tabular}{l | r r r r } 
\hline
& \multicolumn{4}{c}{ASTE}  \\\hline
Model & L14 & R14 & R15 & R16 \\\hline
Paraphrase-T5 & 61.13 & 72.03 & 62.56 & 71.70 \\
Seq2Path & \textbf{64.09} & \textbf{74.29} & \underline{65.42} & \underline{73.67} \\
MvP & \underline{63.33} & \underline{74.05} & 64.53 & 72.76\\\hline
T5 w/ CRO & 62.45& 73.75&\textbf{66.25}&\textbf{74.78}\\\hline
\end{tabular}
\end{table}

\subsection{Contextual Representation Optimization}
As mentioned in Section \ref{Introduce}, some works \cite{yu2023syngen,fei2022lasuie} propose introducing syntactic structure to improve performance in ABSA tasks for generative models. In the Natural Language Processing field, syntax is commonly utilized to refine the representations generated by pre-trained language models. Meanwhile, to fully leverage the insights from both semantic and syntactic channel, sophisticated algorithms for information fusion are essential. Building on previous research, we design a similar approach to optimize contextual representations, which integrates syntactic information to enhance the representations and employs a dynamic gate mechanism \cite{yu2023syngen} to fuse these two channels. As shown in Table \ref{tab:Reprensatation optimization}, we have conducted experiments for T5 model in ASTE task. However, this approach does not always achieve the optimal performance, which indicates a limitation when transitioning from discriminative models to generative ones. One possible explanation is the difference in the pre-training phase: in contrast to independently initialized classifiers, generative models usually contain a pre-trianed decoder. Thus, direct modifications to contextual representation may lead to a semantic gap during auto-regressive generation. Furthermore, since the syntactic structures like dependency tree often contains noisy signals of irrelevant associations, methods that heavily rely on syntax may struggle with accurately aligning nuanced aspects , opinions and sentiments. 

\section{Methodology}\label{Methodology}
In this section, we will first introduce two contextual optimization methods, rule-based approach and score-based approach. Next, based on the score-based optimization method, we further propose PGSO model.
\subsection{Contextual Sequence Optimization Methods}\label{CSOM}
Based on the viewpoint illustrated in the section \ref{Preliminary}, we design two optimization methods (i.e., rule-based static optimization method and score-based dynamic optimization method) to regulate the contextual sequence.
\begin{algorithm}[!h]
    \caption{Processes of the rule-based static optimization method.}
    \label{alg:static sequence}
    \renewcommand{\algorithmicrequire}{\textbf{Input:}}
    \renewcommand{\algorithmicensure}{\textbf{Output:}}
    \begin{algorithmic}[1]
        \REQUIRE The original contextual representations $\emph{H}=\left\{h_i\right\}_{n}$ 
        \ENSURE The contextual representations with optimized sequence ${\emph{G}}=\left\{g_i\right\}_{n}$  
        \STATE $initial(queue)$ $\quad$  //Initial the output queue
        \STATE $initial(sortedQueue)$  $\quad$ //Initial the sort queue
        \STATE //Parsing analysis
        \FOR{$i \in [1,n]$}
            \STATE $Pos(i)= Parsing(w_i)$
        \ENDFOR
        \STATE //Dependency tree $\mathcal{G}(\mathcal{V}, \mathcal{E})$ construction
        \STATE $construct$\, $\mathcal{G}(\mathcal{V}, \mathcal{E})$\, $based$\, $on$\, $Pos(i)$ 
        \STATE $root =getRoot(\mathcal{G})$
        \STATE //Execute Breadth-First Search and output $queue$
        \STATE  $queue.enqueue(root)$
        \WHILE{$queue != empty$}
            \STATE $node = queue.dequeue()$
            \STATE $visit(node$)
            \STATE $children=getChildren(node$)
            \STATE //Sort the children based on the pre-defined rule
            \STATE $sort(children$)
            \FOR{$child \in children$}
            \IF {$child != null$}
                \STATE $sortedQueue.enqueue(child)$
            \ENDIF
            \ENDFOR
            \WHILE{$sortedQueue != empty$}
                \STATE $queue.enqueue(sortedQueue.dequeue())$
            \ENDWHILE
        \ENDWHILE
        \STATE //Regulate sequence based on the queue
        \FOR{$i \in [1,n]$}
            \STATE $g_i = {h}[{queue(i)}]$ 
        \ENDFOR
        \RETURN $\emph{G}$
    \end{algorithmic}
\end{algorithm}

\begin{figure}[ht]
\centering	\includegraphics[width=0.49\textwidth,height=0.72\linewidth]{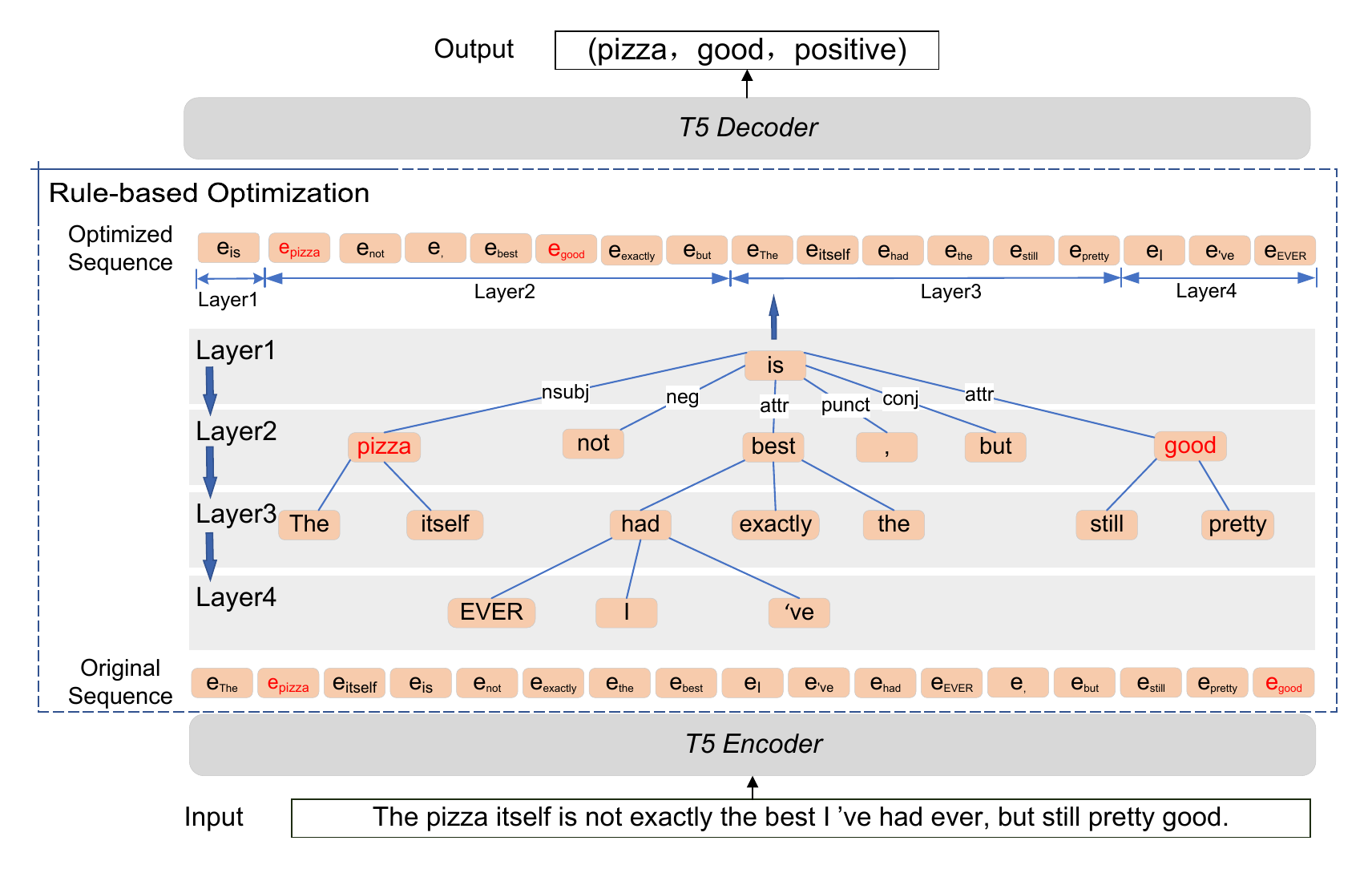}
	\caption{Structure of the rule-based static optimization method. }
	\label{Fig:static sequence}
\end{figure}

\subsubsection{Rule-based Static Optimization}
 We first consider designing a regulating rule by the dependency feature of each word. The processes of the rule-based static optimization method are shown as Algorithm \ref{alg:static sequence} and its structure is shown in Figure \ref{Fig:static sequence}.

The rule-based static optimization method can be streamlined into three processes. (1) \textbf{Parsing analysis and dependency tree construction}: we first execute parsing for the input text by Spacy \footnote{\url{https://spacy.io/}} parsing tool, to construct a dependency tree that represents the grammatical structure of the sentence, which corresponds to lines 3-9 in Algorithm \ref{alg:static sequence}. (2) \textbf{Breadth-First Search and sorting}: we utilize Breadth-First Search (BFS) algorithm traverse the dependency tree and identify the nodes at each layer. These nodes are then sorted according to a pre-defined set of rules, such as the priority of dependency relation. (3) \textbf{Optimizing and output}: based on the sorted node set, we refine the contextual sequence, which is then provided to the decoder for further processing.
\begin{figure}[ht]
\centering	\includegraphics[width=0.49\textwidth,height=0.75\linewidth]{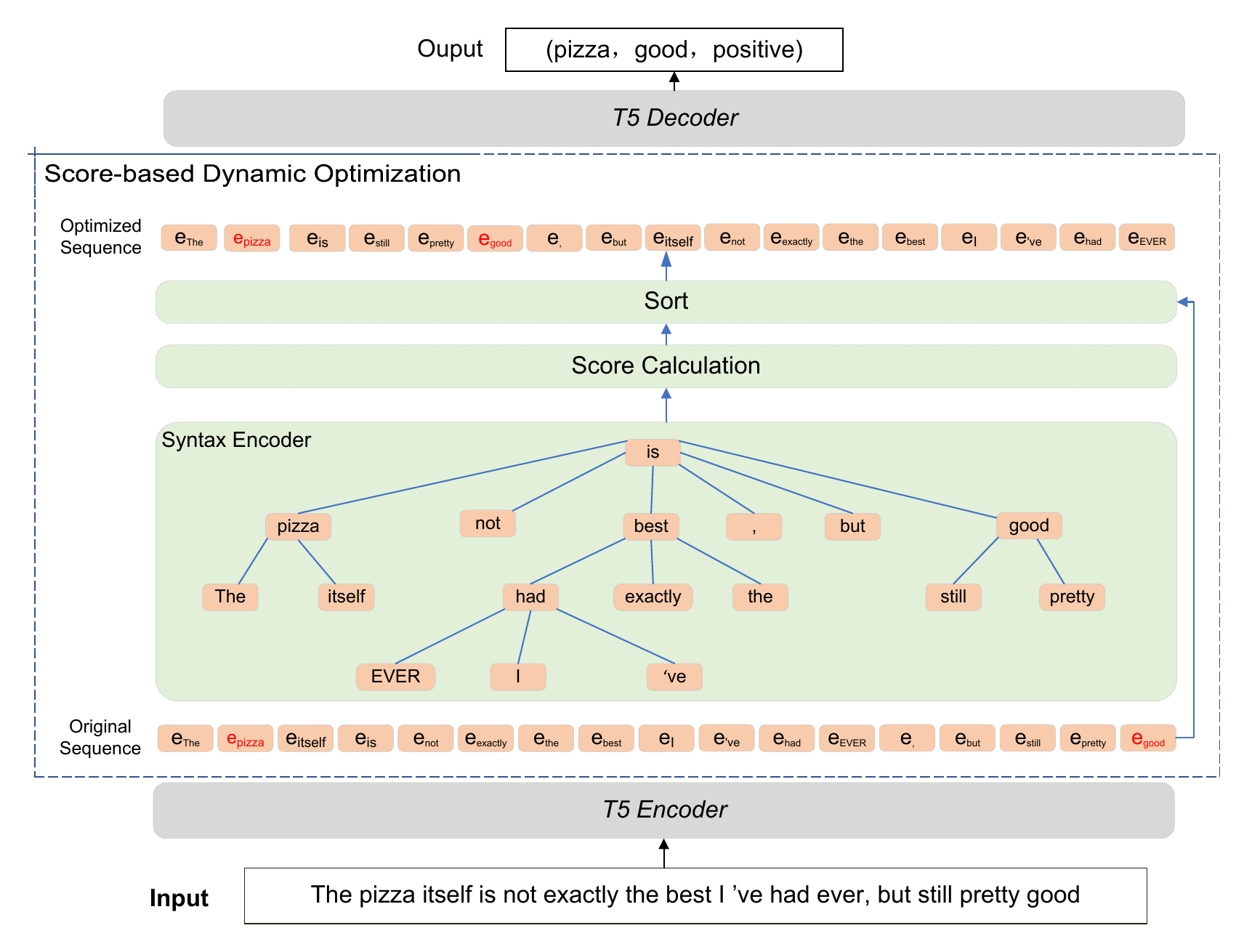}
	\caption{Structure of the score-based dynamic optimization method. }
	\label{Fig:Dynamic Sequence}
\end{figure}
\subsubsection{Score-based Dynamic Optimization}
To further explore syntactic information, we propose a score-based dynamic optimization method, whose structure is shown in Figure \ref{Fig:Dynamic Sequence}. 
The processes of the method can be divided into two steps: (1) \textbf{Representations enhancement}: we first utilize syntax information to enhance the representations from the encoder, whose implementation details will be illustrated in the Syntax Encoder \ref{syntax encoder}. (2) \textbf{Evaluation function}: to mine the semantic and syntactic information, we design a score-based evaluation function, whose implementation details are shown in Score Calculator \ref{score calc}.

Base on the these two optimization structure, we further propose a Prompt-based Generative Sequence Optimization (named PGSO) model to joint optimize the training targets and language model, which will be described in the next Section.

\begin{figure*}[!ht]
\centering	\includegraphics[width=1.\textwidth,height=0.36\linewidth]{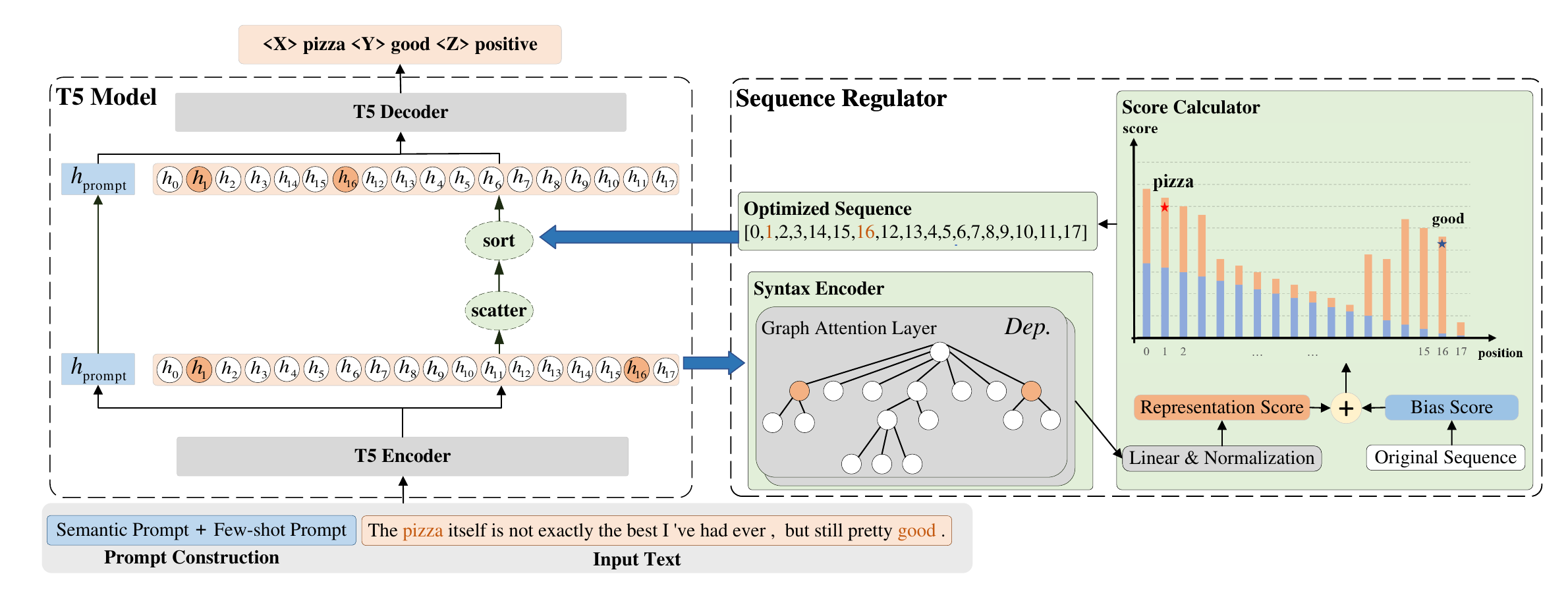}
\caption{Overall architecture of PGSO. The architecture of the Prompt-based Generative Sequence Optimization (PGSO) model extends beyond the conventional encoder-decoder framework of the T5 model to incorporate two distinct components.
\textbf{Prompt Construction}: This component is designed to narrow the gap between pre-training task and downstream ABSA task, maximizing the utilization of our proposed model. It is composed of two specialized prompts: a semantic prompt and a few-shot prompt.
\textbf{Sequence Regulator}: This module includes a syntax encoder and a score calculator. The syntax encoder leverages rich syntax information to enhance the textual representations, thereby enhancing the model's interpretative ability. The score calculator operates on the refined representations to obtain the position score to each word in the input text. Subsequently, it produces the optimized sequence, which is meticulously ordered based on the computed scores, thereby ensuring that the output is not only syntactically coherent but also semantically rich and contextually relevant.}\label{Fig:main structure}
\end{figure*}
\subsection{Overview of PGSO}
As shown in Figure \ref{Fig:main structure}, our proposed model takes the text \emph{s} with task-specific prompt $\emph{s}_{prompt}$ as the input, and outputs the structural sentiment tuples. The architecture of the Prompt-based Generative Sequence Optimization (PGSO) model extends beyond the conventional encoder-decoder framework of the T5 model to incorporate two distinct components.
\textbf{Prompt Construction}: this component is designed to narrow the gap between pre-training task and downstream ABSA task, maximizing the utilization of our proposed model. It is composed of two specialized prompts: a semantic prompt and a few-shot prompt.
\textbf{Sequence Regulator}: as the implementation of the score-based dynamic optimization method, this module includes a syntax encoder and a score calculator. The syntax encoder leverages rich syntax information to enhance the contextual representations, thereby enhancing the model's interpretative ability. The score calculator operates on the refined representations to obtain the position score to each word in the input text. Subsequently, it produces the optimized sequence, which is meticulously ordered based on the computed scores, thereby ensuring that the output is not only syntactically coherent but also semantically rich and contextually relevant.

\begin{table*}[ht]
\centering\caption{Comparisons between T5 unsupervised training and prompt construction for ASTE task in PGSO. In the pre-training phase, the task is to predict randomly masked spans with remaining text. Building upon this principle, we incorporate task-specific prompts in both the input and target to construct a similar style.}\label{tab:Prompt Template}
\scalebox{0.86}{
\begin{tabular}{c | c | c} \hline
\multicolumn{1}{c|}{\textbf{Category}} & \multicolumn{1}{c|}{\textbf{T5 Unsupervised Training}} & \multicolumn{1}{c}{\textbf{Prompt Construction}}\\\hline
Original Text& Thank you for inviting me to your party last week.&The staff is incredibly helpful and attentive.\\\hline
Semantic Prompt	& & Aspect mean \textless X\textgreater, opinion mean \textless Y\textgreater, sentiment mean \textless Z\textgreater. \\\hline
Few-shot Prompt	& &Input : sushi is good. Target : \textless X\textgreater sushi \textless Y\textgreater good \textless Z\textgreater positive. \\\hline\hline
\makecell{Input}& \makecell{Thank you \textless X\textgreater me to your party \textless Y\textgreater week.}& \makecell[c]{Aspect mean \textless X\textgreater, opinion mean \textless Y\textgreater, sentiment mean \textless Z\textgreater.\\Input : sushi is good . 
 Target : \textless X\textgreater sushi \textless Y\textgreater good \textless Z\textgreater positive.\\ The staff is incredibly helpful and attentive.}\\\hline\hline
Target & \textless X\textgreater for inviting \textless Y\textgreater last \textless Z\textgreater	& \makecell{\textless X\textgreater staff \textless Y\textgreater helpful \textless Z\textgreater positive \textless W\textgreater\\  \textless X\textgreater staff \textless Y\textgreater attentive \textless Z\textgreater positive}\\\hline
\end{tabular}
}
\end{table*}


\subsection{Prompt Construction}
Inspired by previous works \cite{gao-etal-2022-lego,gou-etal-2023-mvp}, we adopt prompt-based methods to transfer the sequence generation task to the cloze-style format, aligning more closely with pre-training paradigm. The prompt contains two components: a semantic prompt and a few-shot prompt. 
\subsubsection{Semantic Prompt}
The semantic prompt is constructed with sentiment terms to be predicted in ABSA tasks, accompanied by corresponding sentinel words. As shown in Table \ref{tab:Prompt Template}, the format of the prompt is ``[Sentiment Term] \emph{means} [Sentinel Word]'', adopting the word \emph{means} to connect the sentiment terms and sentinel words, ensuring semantic coherence. Meanwhile, to align with the prompt, the target is also reformulated as a combination of sentiment elements with corresponding sentinel words, whose format is ``[Sentinel Word] \emph{sentiment element}''. Notably, one extra sentinel word will be inserted between sentiment tuples in the target to facilitate model recognition of the start and end of each tuple.
For instance, ASTE task is to extract the sentiment triplet$(a,o,s)$. Thus, in semantic prompt, three sentinel words (\textless X\textgreater, \textless Y\textgreater, \textless Z\textgreater) will be designated for aspect term, opinion term and sentiment term respectively, while one more sentinel word (\textless W\textgreater) will be allocated between the tuples.

\subsubsection{Few-shot Prompt}
To fully utilize the proposed model, we insert a few-shot prompt between the semantic prompt and input text. Specifically, we choose to adopt only an fixed-template artificial one-shot prompt for all tasks. Despite multiple shots may provide improvement, the risk of performance decline is associated with improperly crafted prompt cases due to the high sensitivity of the pre-trained language model to prompts. Importantly, our primary emphasis in the proposed model is on advancing sequence optimization rather than the prompt design.

\subsection{Sequence Regulator}
In this section, we will introduce the Sequence Regulator, the key module of PGSO. It takes the representations from encoder and output the optimized sequence to re-rank the context. As shown in Figure \ref{Fig:main structure}, there are two components: the syntax encoder and the sequence regulator.

\subsubsection{Syntax Encoder}\label{syntax encoder}
The text representations from T5 encoder primarily contains semantic information. To harness rich syntactic details, we introduce a syntax encoder. Specifically, syntax encoder module utilizes a graph attention network (GAT) \cite{velivckovic2017graph} composed by multiple graph attention layers guided by the dependency tree. The dependency tree is considered as a directed graph, whose adjacent matrix $DA$ is formulated in the Equation \ref{DA} and GAT processes can be formulated in the Equation \ref{fuse}, \ref{norm} and \ref{active}.
\begin{equation}\label{DA}\centering
{DA}_{i, j} = 
\begin{cases}
    1 & \text { if } w_i \text{ is the parent of }w_j \text{ in Dep.Tree} \\
    0 & \text { otherwise }
\end{cases}
\end{equation}
\begin{equation}\label{fuse}
    g_i^{l+1}=\|_{k=1}^K \sum_{j \in \mathcal{N}_i} \alpha_{i j}^{l k} \mathbf{W}_k^l g_j^l   
\end{equation}
\begin{equation}\label{norm}
    \alpha_{i j}^{l k}=\frac{\exp \left(e_{i j}^{l k}\right)}{\sum_{j=1}^{\mathcal{N}_i} \exp \left(e_{i j}^{l k}\right)}   
\end{equation}
\begin{equation}\label{active}
    e_{i j}^{l k}=\operatorname{LeakyReLU}\left({{\mathbf{a}_{\mathbf{}}}}^T\left[\mathbf{W}^{l k} g_{i}^{l} \| \mathbf{W}_{\mathbf{}}^{l k} g_{i}^{l}\right]\right)
\end{equation}
where $\mathcal{N}_i$ is the set of neighbors of $w_i$, $g_i^{l}$ is the representation of $w_i$ in layer $l$, $||$ denotes vector concatenation, $K$ is the number of attention heads, $\mathbf{W}^{l k}$, $\mathbf{a}$ are trainable parameters of the $k$th head of layer $l$, $\operatorname{LeakyReLU}$ is the activation function. The initial representations $g^0=h$, where the $h$ stands for the contextual representations, and the final output representations from the syntax encoder is $g$. 

\subsubsection{Score Calculator}\label{score calc}
To transfer the contextual sequence as a trainable variable, we realize the score-based evaluation function, which jointly considers the semantic, syntactic and original sequence information. Specifically, we introduce position score for each word to quantify its positional significance, which is composed by two parts: the representation score and the bias score, which is formulated as follows,
\begin{equation}
    s_i^{ps} = s_{i}^{rs}+s_{i}^{bs}
\end{equation}
where $s_i^{ps}$, $s_{i}^{rs}$, $s_{i}^{bs}$ are the position score, representation score and bias score of the word $w_i$ respectively.

\textbf{Representation Score}: 
To obtain the latent optimal decoding sequence, we design a unified approach to leverage the representations from the syntax encoder. It consists a Linear layer and Normalization function, which can be formulated as follows,
\begin{equation}
    s_{i}^{rs}=\frac{\exp \left(\mathbf{W}g_i\right)}{\sum_{j=1}^{n} \exp \left(\mathbf{W}g_j\right)}  
\end{equation}

\textbf{Bias Score}: 
Given that attention calculation is order-independent, the rearrangement scheme relying solely on representation score may lead to over-adjustment due to syntactic noise propagation, especially for the short texts. To mitigate the issue, we also introduce bias score, explicitly leveraging the original sequence to provide a hierarchical rectified gradient. Specifically, for long texts, low-gradient preserves free-adjustment features, facilitating modeling long-distance relations. Conversely, in short texts, high-gradient provides resistance to rectify the over-adjustment in short texts. The calculation processes of the bias score are as follows,   

\begin{equation}  
    s_{i}^{bs} = \frac{\exp \left(l_i\right)}{\sum_{j=1}^{n} \exp \left(l_j\right)}
\end{equation}

\begin{equation}\label{bias}
    {l_i} = \frac{l-i\times d}{n}
\end{equation}
where $n$ is the length of text. $l$ and $d$ are pre-defined hyper-parameters length and step respectively.
For a clearer comprehension of the function of the proposed bias score, we consider examples with text lengths of 3 and 18. As shown in Table \ref{tab:score bais}, it is noteworthy that the initial interval for short texts is substantially larger than that for long texts, ensuring the positions are kept relatively stable. This observation underscores the resistance to over-adjustment issue, particularly in the context of short texts.
\begin{table*}[pos=!ht]
\centering\caption{Data statistics of the mentioned datasets.}\label{tab:Data Statics}\scalebox{0.9}{
\begin{tabular}{c c | c c c c | c c | c c c c| c c} 
\hline
& &\multicolumn{4}{c|}{ASTE} & \multicolumn{2}{c|}
{TASD}&\multicolumn{4}{c|}{UABSA} & \multicolumn{2}{c}{ACOS}\\
  Set& Sentiment Polarity&L14 & R14 & R15 & R16 & R15 & R16 & L14 & R14 & R15 & R16 & L14 & R16\\\hline
 Training & Positive & 817&	1692&	783&	1015&	1198&	1657&	882&	1957&	812&	1119&	2583&	1656\\
 & Neutral & 126&	166&	25&	50&53&	101&	408&	575&	34&	60&	227&	95\\
& Negative & 517&	480&	205&	329&	403&	749&	755&	737&	233&	410&	1364&	733\\\hline
Validation & Positive & 169& 404&185&252&6&23&104& 213& 102&122&279&180 \\
& Neutral & 36& 54&11&11&0&1& 46& 53&2&9&24 & 12\\
& Negative & 141& 119&53&76&7&20& 106& 64&26&36&137 & 69\\\hline
 Testing & Positive & 364& 773&317&407&454&611&339&728&327&471&716 &668\\
 & Neutral & 63& 66&25&29&45&44& 165& 198&34&32&65 & 44\\
& Negative & 116& 155&143&78&346&204&130&195&186&119&380& 205\\\hline
\end{tabular}}
\end{table*} 

\begin{table}[ht]
\centering{\caption{The distribution of position scores $s^{bs}$ under varying text lengths $n$ is presented. Notably, starting from the initial interval, it is noteworthy that the interval for short texts is substantially larger than that for long texts. This observation underscores the resistance to over-adjustment issue, particularly in the context of short texts.}\label{tab:score bais}
\scalebox{0.7}{
\begin{tabular}{c|c c c c c c |c } \hline
{Text Length} & \multicolumn{6}{c|}{Position}&{Interval} \\\hline
& 0&1&2&3&...&17&{0}$\sim${1} \\\hline
$n$=3 & 0.4484&0.3213&0.2302& & & & 0.1271\\
$n$=18 & 0.0855&0.0809&0.0765&0.00724 &... & 0.00332& 0.0046 \\\hline
\end{tabular}
}
}
\end{table}

\section{Experiment}\label{Experiment}
\subsection{Datasets}
We evaluate the PGSO model on 12 datasets over four tasks, including Laptop14, Rest14, Rest15 and Rest16. These 12 datasets are originally provided by the SemEval shared challenges \cite{pontiki-etal-2014-semeval,pontiki-etal-2015-semeval,pontiki-etal-2016-semeval}. Specifically, for ASTE, TASD and UABSA tasks, we adopt the datasets provided by \cite{xu-etal-2020-position,DBLP:conf/aaai/WanYDLQP20,Li_Bing_Li_Lam_2019}. For ACOS tasks, the dataset is provided by \cite{DBLP:conf/aaai/WanYDLQP20}. The data statistics are shown in Table \ref{tab:Data Statics}.

All results are the average F1 scores across three runs with different random seeds.To align with the settings of previous works, we adopt T5-base model from Huggingface Transformers Library \footnote{\url{https://github.com/huggingface/transformers}} as our backbone model. The learning rate is set to be $3\times 10^{-4}$ and training epoch is set to 40. In the syntax encoder module, the number of graph attention layers is 2, the dropout rate is 0.4 and the alpha is 0.05. To align with the T5-base model, the $l$ and $d$ in bias score are fixed to 128 and 1 respectively. The details of each hyper-parameters are listed in the appendix C.

\subsection{Baselines}
We compare our model with discrimination-based methods and generation-based methods, which are introduced as follows:

(1)	Discrimination-based methods:
\textbf{Span-ASTE} \cite{xu-etal-2021-learning} proposes a Span-BERT based methods to learn interactions between target spans and opinion spans for the ASTE task. \textbf{SBN} \cite{chen-etal-2022-span} is another span-level bidirectional network for ASTE task.
\textbf{RACL} \cite{chen-qian-2020-relation} proposes a relation propagation mechanisms to tackle ABSA tasks based on BERT model. \textbf{Jet-BERT} \cite{xu-etal-2020-position} tackles ABSA tasks in an end-to-end manner by a tagging scheme. \textbf{Dual-MRC} \cite{Mao_Shen_Yu_Cai_2021} is a dual-channel MRC (Machine Reading Comprehension) structure to tackle triplet extraction task. Similarly, \textbf{BMRC} \cite{Chen_Wang_Liu_Wang_2021} is a Bi-direction MRC model to extract aspect and opinion separately. \textbf{Extract-Classify-ACOS} \cite{cai-etal-2021-aspect} is the first work to propose ACOS task, and propose a two-step structure based on BERT model.

(2) Generation-based methods: \textbf{GAS-T5} \cite{zhang-etal-2021-towards-generative} is the first work to adopt T5 as backbone model to address ABSA tasks. \textbf{Paraphrase-T5} \cite{zhang-etal-2021-aspect-sentiment} proposes a paraphrasing template to exploit semantic relation between sentiment elements. \textbf{Seq2Path} \cite{mao-etal-2022-seq2path} treats sentiment tuple as a path of the tree, predicting tuples separately. \textbf{DLO} \cite{hu-etal-2022-improving-aspect} investigates the order of sentiment elements. \textbf{LEGO-ABSA} \cite{gao-etal-2022-lego} proposes a LEGO-style prompt assemble structure. \textbf{MvP} \cite{gou-etal-2023-mvp} aggregates sentiment elements generated in different orders. \textbf{BARTABSA} \cite{10013820} proposes a BART based model to tackle ACOS task. \textbf{BART-CRN} \cite{xiong2023bart} tackles ACOS extraction as a sequence generation task. \textbf{EHG} \cite{lv-etal-2023-efficient} leverages an Efficient Hybrid Transformer to generate relations.
 
\subsection{Overall Performance}
We have conducted extensive experiments on 4 tasks over 12 datasets. The overall performance comparison is shown in Table \ref{tab:ASTE result}, \ref{tab:TASD result}, \ref{tab:UABSA result} and \ref{tab:ACOS result}. Most baselines are copied from \cite{mao-etal-2022-seq2path}. Our proposed model obtains the state-of-the-art results in almost all F1 scores. 
\begin{table}[pos=h]
\centering{\caption{Overall performance of the ASTE task over 4 datasets. Note that the best results and results with F1 gaps within 0.03 are in \textbf{bold}, and second-best are \underline{underline}.}\label{tab:ASTE result}
\begin{tabular}{l c c c c c} 
\\\hline
 Model  &L14 & R14 & R15 & R16 & AVG\\\hline
Jet-BERT & 51.04 & 62.40 & 57.53 & 63.83 & 58.70\\
Dual-MRC & 55.58 & 70.32 & 57.21 & 67.40  &62.62\\
BMRC & 59.27 & 70.69 & 61.05  & 68.13  & 64.78\\
Span-ASTE & 59.38 &71.85 &63.27 &70.26 & 66.19\\
SBN &62.65 & \underline{74.34} & 64.82 &72.08 & 68.47 \\
GAS-T5 & 60.78 & 72.16 & 62.10 & 70.10 & 66.28\\
ParaPhrase-T5 & 61.13 & 72.03 & 62.56 & 71.70 & 66.85\\
Seq2Path & \underline{64.09} & 74.29 & 65.42 & 73.67 & \underline{69.36}\\
DLO & 61.46 & 72.39 & 64.26 & 69.90 & 67.00\\
LEGO-ABSA & 62.22 & 73.21 & 64.46 & 71.59 & 67.87\\
EHG & 61.53 & 71.82 & 63.58 & 72.35&67.32\\
MvP & 63.33 & 74.05 &65.89 & 73.48&69.18\\\hline
$PGSO_{static}$ & 62.93 & 73.86 & \underline{66.13}& \underline{74.11}& 69.25\\
$PGSO_{dynamic}$ & \textbf{64.14}& \textbf{74.38}&\textbf{67.28}&\textbf{75.33}& \textbf{70.28}\\\hline
\end{tabular}
}
\end{table}

On the ASTE task, our proposed PGSO model based on dynamic optimization method achieves a 0.92 improvement in F1 score over the previous best result. While the performance of our static rule-based PGSO is marginally below that of Seq2Path, this suggests that there is room for further refinement in the formulation of our rules. As one of the most widespread tasks, the challenge of ASTE task lies in accurately modeling the relations between aspects and opinions, and making correct sentiment predictions. Our proposed PGSO exhibits superior ability in relation extraction.
\begin{table}[pos=!ht]
\centering{\caption{Main results of the TASD task over two datasets. Note that the best results and results with F1 gaps within 0.03 are in \textbf{bold}, and second-best are \underline{underline}.}\label{tab:TASD result}
\begin{tabular}{l c c c c} 
\\\hline
 Model &  R15 & R16  & AVG \\\hline
GAS-T5 & 61.47 &69.42 &65.44 \\
ParaPhrase-T5& 63.06 & 71.97& 67.51\\
Seq2Path & 63.13 & 68.47 & 65.80\\
DLO & 62.95 & 71.79 & 67.37\\
LEGO-ABSA & 63.15 & 72.02 & 67.58\\
MvP & \underline{64.53} & \textbf{72.76} & \underline{68.64} \\\hline
$PGSO_{static}$ & \underline{65.09} & 71.86 & 68.47 \\
$PGSO_{dynamic}$ &\textbf{65.40}&\textbf{72.74} &\textbf{69.07}\\\hline
\end{tabular}
}
\end{table}
\begin{table}[pos=!ht]
\centering{\caption{Main results of the UABSA task over 4 datasets. Note that the best results and results with F1 gaps within 0.03 are in \textbf{bold}, and second-best are \underline{underline}.}\label{tab:UABSA result}
\begin{tabular}{l  c c c c c} 
\\\hline
 Model & L14 & R14 & R15 & R16 &AVG \\\hline
 RACL & 63.40 &75.42 &66.05 &-&-\\
Dual-MRC& 65.94 & 75.95 & 65.08 & - &-\\
BMRC &  67.27  & 76.39  & 67.16  & 73.18 &  71.00 \\
GAS-T5 &68.64& 77.13 & 66.78&73.64 & 71.54\\
Seq2Path& 70.00 & 77.01 & 68.35& 75.87 & 72.80\\\hline
$PGSO_{static}$ & \underline{71.33}& \underline{78.26}&\underline{69.21}&\underline{75.98}&\underline{73.69}\\
$PGSO_{dynamic}$ & \textbf{72.28}& \textbf{78.38}&\textbf{70.76}&\textbf{76.55}&\textbf{74.49}\\\hline
\end{tabular}
}
\end{table}
\begin{table}[pos=!ht]
\centering{\caption{Main results of the ACOS tasks over two datasets. Note that the best results and results with F1 gaps within 0.03 are in \textbf{bold}, and second-best are \underline{underline}.}\label{tab:ACOS result}
\begin{tabular}{l  c c c} 
\\\hline
Model & L14 & R16 & AVG\\\hline
Extract-Classify-ACOS & 35.80&44.61&40.20\\
BARTABSA &39.41 & 53.45 & 46.43\\
BART-CRN & 38.32 & 48.90 & 43.61\\
Seq2Path & 42.97&58.41 & 50.69 \\
DLO & 43.64& 59.99 & 51.81\\ 
MvP & 43.92& \textbf{61.54} & \underline{52.73}\\\hline
$PGSO_{static}$&\underline{44.53} & 60.86 & 52.69\\
$PGSO_{dynamic}$&\textbf{44.77} & \textbf{61.51} & \textbf{53.14}\\\hline
\end{tabular}
}
\end{table}
\begin{table*}[pos=!ht]
\centering\caption{Ablation Study. Notations "PC" represents \textbf{P}rompt \textbf{C}onstruction, "SR" represents \textbf{S}equence \textbf{R}egulator, "SP" represents \textbf{S}emantic \textbf{P}rompt in prompt construction, "FP" represents \textbf{F}ew-shot \textbf{P}rompt in prompt construction, "BS" represents the \textbf{B}ias \textbf{S}core in sequence regulator. Note that the best results and results with F1 gaps within 0.03 are in \textbf{bold}, and second-best are \underline{underline}.}\label{tab:Ablation Study}\scalebox{0.9}{
\begin{tabular}{l l | c c c c | c c | c c c c| c c |c} 
\hline
&& \multicolumn{4}{c|}{ASTE} & \multicolumn{2}{c|}
{TASD}&\multicolumn{4}{c|}{UABSA} & \multicolumn{2}{c|}{ACOS}&\\
 Category&Ablation& L14 & R14 & R15 & R16 & R15 & R16 & L14 & R14 & R15 & R16 & L14 & R16&AVG\\\hline
   w/o SR & w/o PC& 62.32& 72.55&62.87&70.15&62.57&69.70&68.75&76.95&67.80&73.75&44.17 &56.90&65.71\\
 & w/ FP & 61.97& 72.85&62.70&71.17&63.33&69.91&69.01&77.11&68.23&74.15&44.18 &57.10&65.98\\
 & w/ SP& 63.10& 73.11&65.87&72.90&64.37&72.19&71.69&78.20&68.91&75.71&44.21 &61.10&67.61\\
\hline\hline
w/o PC & w/o SR& 62.32& 72.55&62.87&70.15&62.57&69.70&68.75&76.95&67.80&73.75&44.17 &56.90&65.71\\
& w/ BS & 63.03& \underline{73.39}&64.21&71.61&62.71&70.35& 71.33& 78.24&68.93&73.99&44.21 & 57.79&66.65\\\hline
w/ PC& w/o SR & 62.35& 73.25&66.62&73.96&64.39&\underline{72.68}&71.33&78.32&69.81&76.11&44.42& \underline{61.25}&67.87\\
& w/o BS& \underline{64.10}& 73.06&\underline{66.80}&\underline{74.28}&\underline{65.21}&71.93& \underline{72.04}& \textbf{78.37}&\textbf{70.78}&\underline{76.34}&\underline{44.60} &60.31&\underline{68.15}\\
& w/ BS& \textbf{64.14}& \textbf{74.38}&\textbf{67.28}&\textbf{75.33}&\textbf{65.40}&\textbf{72.74}& \textbf{72.28}& \textbf{78.38}&\textbf{70.76}&\textbf{76.55}&\textbf{44.77} & \textbf{61.51}&\textbf{68.63}\\
\hline
\end{tabular}}
\end{table*}
On the TASD task, our proposed PGSO model based on the dynamic regulating method method a 0.43 improvement in F1 score over the previous best result. Compared to ASTE task, since opinion term is not the element to be extracted in TASD task, which may results in less significant performance gains for the PGSO model.

On the UABSA task, our proposed model based on the dynamic regulating method and static rule-based regulating method achieve 1.69 and 0.89 improvement in F1 score over the previous best result respectively. Since the UABSA task is just to predict the aspect and its corresponding sentiment, the prompt-based methods can effectively model the relation between the sentiment elements. 

On the ACOS task, our proposed model based on the dynamic regulating method achieves 0.41 improvement in F1 score over the previous best result. Compared with discriminative manners, generative approaches exhibit overall superior performance due to the complex relations in the quadruple extraction task.

\subsection{Ablation Study}
We also conduct an ablation study to verify each component's effectiveness in our proposed PGSO based on the dynamic optimization method. The results are shown in Table \ref{tab:Ablation Study}, and the observations are as follows.

(1) The original T5 model achieves the lowest result, indicating that the pre-trained language model is not effectively utilized without any optimization methods. 

(2) To verify the impact of different types of prompts, we conducted additional experiments varying the prompt categories, focusing on the effectiveness of semantic prompts and few-shot prompts. In category \emph{w/o SR}, the model equipped with semantic prompts outperformed the baseline T5 model, achieving an average enhancement of 1.90 in F1 score. This significant improvement suggests that semantic prompts effectively convert the original generation task into a cloze-style task, which aligns well with the model's pre-training objectives. In contrast, the few-shot prompt only resulted in a modest 0.27 increase in F1 score, indicating that the model's performance is indeed sensitive to the nature of the prompts used.

(3) Within category \emph{w/o PC}, \textbf{w/o SR} performs inferior to \textbf{w/ SR}, which implies that the dynamic optimization scheme is also effective even without any prompts. 

(4) We also conduct experiment to identify the effectiveness of representations score. In the category \emph{w/ PC}, compared with \textbf{w/o SR}, \textbf{w/o BS} achieves superior results except TASD and ACOS tasks under Rest16 dataset. This phenomenon suggests that while algorithms relying on representation scores are generally effective, they sometimes overlook the information contained in the original sequence, which can result in incorrect predictions in certain instances. 

(4) To verify the function of bias score in the sequence regulator, we also conduct comparison experiment. In category \emph{w/ PC}, \textbf{w/ BS} is superior to \textbf{w/o BS} on almost the datasets, which means that introducing original sequence can boost performance, rectifying the aforementioned incorrect predictions.

\section{Analysis}\label{Analysis}
\subsection{Performance in different relational distance}
To show the distinctive superiority of our proposed PGSO in long-distance relation extraction, we conduct experiments under different distance relations. 
\begin{figure}[!ht]
	\centering	\includegraphics[width=0.49\textwidth,height=0.75\linewidth]{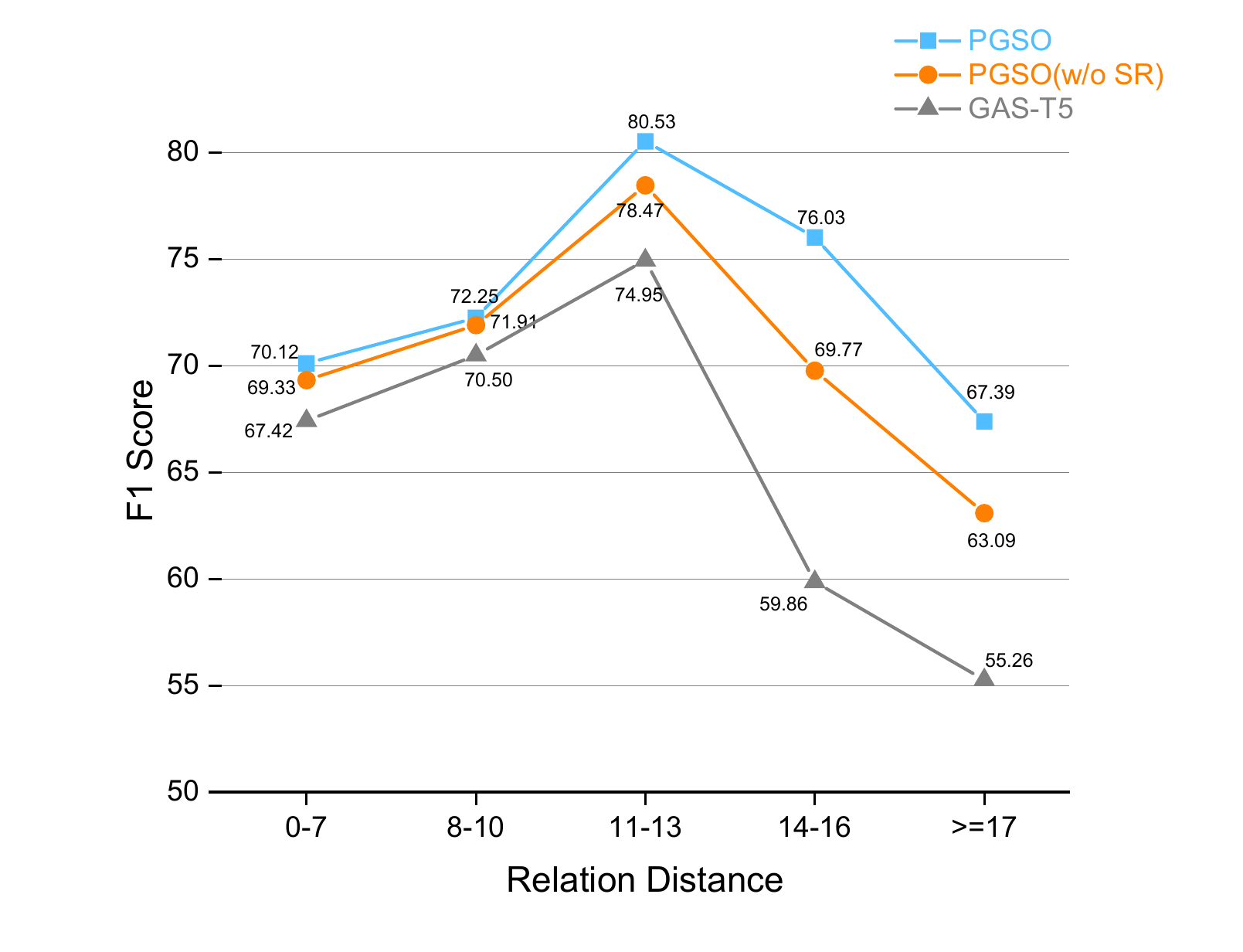}
	\caption{Comparisons of GAS-T5, PGSO (w/o SR) and PGSO with respect to the performace across various distance between the aspect and opinion. Notation "SR" represents \textbf{S}equence \textbf{R}egulator.}
	\label{Fig:Distance F1}
\end{figure}
All the results are average F1 scores across four datasets with three different random seeds in ASTE task. The results are shown in Figure \ref{Fig:Distance F1}. Key observations are as follows.

(1) All models exhibit comparable performance on relationships with a distance of less than 11, with the maximum discrepancy between any two models being less than \textbf{2.7}. This is expected since predicting short relations is relatively straightforward for a model like T5-base, which boasts 220 million parameters.

(2) In the domain of long-distance relations (11 to 16), our PGSO model significantly surpasses its competitors. Specifically, it achieves an F1 score improvement of \textbf{4.17} and \textbf{10.87} over the other two models, indicating the efficacy of our unique sequence rearrangement approach.

(3) Due to the scarcity of extremely long-distance relations, which aligns with the observed long-tail distribution, all models face lower performance results. Despite this challenge, our model stands out, delivering a notable F1 score improvement of \textbf{4.30} and \textbf{12.13} over its competitors.
\begin{table*}[pos=!ht]
\centering\caption{Experiments with different sequence optimization methods. "SR" represents \textbf{S}equence \textbf{R}egulator. Note that the best results and results with F1 gaps within 0.03 are in \textbf{bold}, and second-best are \underline{underline}.}\label{tab:Effects optimization}
{
\begin{tabular}{ l l | c c c c | c c | c c| c} 
\hline
&& \multicolumn{4}{c|}{ASTE} & \multicolumn{2}{c|}
{TASD} & \multicolumn{2}{c|}{ACOS}& \multicolumn{1}{c}{AVG}\\
 Category&Ablation& L14 & R14 & R15 & R16 & R15 & R16 & L14 & R16\\\hline
 w/o SR&- & 62.35&73.25&66.62&73.96&64.39&72.68&44.42&\underline{61.25}&64.86 \\\hline
Rule-based & Rule-1 &61.46& 72.59&66.93&73.69&65.30&70.98& 43.69& 60.96&64.45\\
& Rule-2 & \underline{63.46}&73.37&\underline{67.25}&73.63&64.40&\underline{71.98}&44.53&60.93&64.94\\
& Rule-3 & 62.93& \underline{73.86}&67.13&\underline{74.11}&\underline{65.09}&71.86&\underline{44.59}&60.86&\underline{65.05}\\\hline
Score-based &-& \textbf{64.14}& \textbf{74.38}& \textbf{67.28}&\textbf{75.33}&\textbf{65.40}&\textbf{72.74}&\textbf{44.77}& \textbf{61.51}& \textbf{65.69}\\\hline
\end{tabular}
}
\end{table*}

\begin{table*}[pos=!ht]
\centering\caption{Prediction comparisons from \emph{GAS-T5} and \emph{PGSO}. For each sentence, the aspects and opinions are displayed in \textbf{bold}. False predictions are marked with \XSolidBrush  while true predictions are marked with \Checkmark. POS, NEG, NEU represent positive, negative and 
 neutral sentiment polarities respectively.}\label{Table:Examples}
	\begin{tabular}{c | c c}
            \hline
		Sentences & GAS-T5 & PGSO \\\hline
    \makecell{My mom originally introduced me to this place,\\ but even she (being Indian) feels the \textbf{food} \\can be somewhat over the top \textbf{spicy} and far too \textbf{oily} .}  &\makecell{(food, spicy, NEG)\XSolidBrush}&  \makecell{(food, spicy, NEG), \\(food, oily, NEG)\Checkmark} \\\hline
  \makecell{Bring your cell phone cause you may have to \\wait to get into the \textbf{best} sushi restaurant \\in the world: \textbf{BLUE RIBBON SUSHI} .} & \makecell{(sushi, best, POS)\XSolidBrush}&  \makecell{(BLUE RIBBON SUSHI,\\ best, POS)\Checkmark} \\\hline
  \makecell{The \textbf{pizza} itself is not exactly the best \\I 've had EVER , but still pretty \textbf{good} .}&\makecell{(pizza, best, POS)\XSolidBrush} &  \makecell{(pizza, good, POS)\Checkmark} \\\hline
  \makecell{The \textbf{scallops} are apparently cooked in a \\black olive butter which really \\makes them \textbf{unique}(not to mention \textbf{tasty} ) .}&\makecell{(black olive butter, unique, POS), \\(black olive butter, tasty, POS)\XSolidBrush} &   \makecell{(scallops, unique, POS), \\(scallops, tasty, POS)\Checkmark}\\\hline
	\end{tabular}
\end{table*}

\subsection{Effects of sequence optimization methods}
We also investigate the effects of our sequence optimization methods. In the Section \ref{CSOM}, we introduce a rule-based static optimization and a score-based dynamic optimization method. Firstly, we crafted three distinct sets of rules leveraging the dependency attributes of words, with the detailed rules provided in the Appendix \ref{rule definition}. Secondly, based on the PGSO structure, we experimented with diverse sequence optimization methodologies as potential substitutes for the sequence regulator component.To isolate the effects of different optimization methods, we conducted controlled experiments: one without any sequence optimization and another with a score-based dynamic sequence regulator. The experiment results are shown as Table \ref{tab:Effects optimization}. Key observations are as follows:

(1) Intentionally reversing the dependency ranking led to the model based on rule-1 showing the lowest F1 score of 64.45, which is even lower than the model without any sequence optimization. This suggests that introducing an irrational rule can disrupt the original contextual sequence, negatively affecting the decoder's modeling capabilities. 

(2) Models with well-crafted rules, namely rule-2 and rule-3, achieved sub-optimal results across various ABSA tasks. Compared to the baseline, these models respectively improved their F1 scores by 0.08 and 0.19, demonstrating the effectiveness of sequence optimization algorithms. However, they not surpass the performance of the model with a score-based dynamic sequence optimization method, suggesting that the rule-based static approaches may lack generality.

(3) The model with score-based dynamic optimization method achieves the best result across all the datasets and tasks, which exhibits greater completeness.

\begin{table*}[pos=!ht]
\centering\caption{Complexity analysis. "PC" represents  \textbf{P}rompt \textbf{C}onstruction, "SR" represents \textbf{S}equence \textbf{R}egulator. Note that the reported result reflects the average runtime in the training phrase of two distinct executions. The unit of each time is second.}\label{tab:complexity}
{
\begin{tabular}{ l | c c c c  | c c | c c } 
\hline
 & \multicolumn{4}{c|}{ASTE} & \multicolumn{2}{c|}{TASD} & \multicolumn{2}{c}{AVG}\\
 Model & L14 & R14 & R15 & R16 & R15 & R16& &\\\hline
 PGSO (w/o PC, w/o SR)& 265&385&198&251&299&425 & 303&\\
PGSO (w/ PC, w/o SR) &275&401&204&279&308& 440 & 317 & +4.6\%\\
PGSO (w/ PC, w/ SR) &351& 511&274&372&388&548& 407 & +28.3\%\\\hline
\end{tabular}
}
\end{table*}
\subsection{Case Study}
As shown in Table \ref{Table:Examples}, we present four cases to provide a comprehensive understanding of our proposed model. We choose GAS-T5 \cite{zhang-etal-2021-towards-generative} for comparison, which is a target-oriented optimization method based on the original T5-base model.

The first three examples are typical sentences with long-distance aspect-opinion pairs. While GAS-T5 struggles to extract these relations, PGSO accurately predicts them, demonstrating that our contextual re-ranking mechanism enhances the model's capacity to extract such long-distance aspect-opinion relations. 

The final example features a more complex sentence with an aspect interference term (``\emph{black olive butter}''). Traditionally, the original language model biases attention weights towards closer word pairs, leading to redundant predictions like \emph{(black olive butter, unique, POS)} and \emph{(black olive butter, tasty, POS)} for the GAS-T5 model. In contrast, trough contextual rearrangement, PGSO successfully captures the correct associations and diminishes the interference's impact.
\begin{table*}[pos=!ht]
\centering\caption{Error analysis for our proposed PGSO model. POS, NEG, NEU represent positive, negative and 
 neutral sentiment polarities respectively.}\label{Table:Error}
	\begin{tabular}{c | c | c}
            \hline
		Sentences  & Gold label & Prediction \\\hline
    \makecell{MS Office 2011 for Mac is\\ wonderful, well worth it .}  &\makecell{(MS Office 2011 for Mac,\\ wonderful, POS)} &\makecell{(Mac, wonderful, POS)} \\\hline
  \makecell{excellent food, nice ambiance, fairly expensive} & \makecell{(NULL, RESTAURANT\#PRICES,\\ NEG, expensive)}& \makecell{(food, RESTAURANT\#PRICES,\\ NEG, expensive)} \\\hline
	\end{tabular}
\end{table*}
\subsection{Complexity Analysis}
To assess the impact of our prompt construction and sequence regulator module, we conduct a complexity analysis focusing on two key aspects: model scale and training duration. 

(1) Our sequence regulator module introduces an additional 461,000 parameters, primarily due to the inclusion of Linear layers. This increase is relatively minor when compared to the original T5 model, which already possesses 222 million parameters. This suggests that our approach does not significantly expand the model's size.

(2) Compared to the original T5 model, the model that incorporates the prompt construction module only experiences a 4.6\% increase in training time. This result suggests that adding prompt texts does not substantially extend the model's runtime.

(3) On average, the sequence regulator module demands 28.3\% more training time than a model that solely incorporates the prompt construction module. This increased time consumption could be attributed to two primary factors. Firstly, due to the involvement of tensor scatter and sort operations, the sequence regulation process may not be sufficiently parallelized, which could limit performance. Secondly, our reconstructed generation function might not be as well-optimized as the original one, potentially leading to higher computational costs.

\subsection{Error Analysis}
To conduct comprehensive investigation of our method, we have chosen two typical wrong predictions for in-depth error analysis, aiming to specify the potential direction for the improvement or refinement. The examples are presented in Table \ref{Table:Error}.

The first case pertains to the ambiguity in determining the boundaries of aspect or opinion spans. Despite the model's success in identifying the relation between aspects and opinions, it occasionally fails in predicting the exact spans. Thus, improving the precision of span boundary is a possible future enhancement.

For the second case, in the Aspect-Category-Opinion-Sentiment (ACOS) task, predicting the "NULL" aspect is particularly challenging. One possible reason is the absence of the "NULL" node in the dependency tree. In the syntax encoder module, we introduce a Graph Attention Network (GAT) to capture the relations within the dependency tree. However, when the aspect term is not explicitly mentioned in the text, the aspect-opinion relation is not reflected in the dependency tree, leading to a incorrect prediction. A potential solution is to integrate a "NULL" as a leaf node in the dependency tree, enabling the model to explicitly capture relations.

\section{Conclusion}\label{Conclusion}
In this paper, we first propose two sequence optimization methods to address the limitation of the position embedding mechanism in the PLMs. Based on the score-based dynamic optimization structure, we further propose PGSO, a unified Prompt-based Generative Sequence Optimization network, to boost the long-distance relation extraction by rearranging context. This is the first work to
introduce a model-oriented optimization methods aimed at addressing the limitations of generative models in long-distance relation extraction within ABSA tasks.
Specifically, PGSO contains two components, namely, \emph{prompt construction} method and \emph{sequence regulator} module. 
The former constructs a task-specific prompt based on pre-training objectives, effectively bridging the gap between pre-training and downstream tasks, maximizing utility of the proposed model. The latter adopts syntactic information to dynamically optimize the contextual sequence, thus enhancing the model's ability to identify long-distance relations. Moreover, we have conducted extensive experiments on four ABSA tasks across multiple benchmarks, which demonstrates that PGSO outperforms state-of-the-art methods.
\section{Acknowledge}
This work was supported in part by the National Natural Science Foundation of China under Grant No.62276110, Grant No.61602197, Grant No.61772076, in part by CCF-AFSG Research Fund under Grant No.RF20210005, and in part by the fund of Joint Laboratory of HUST and Pingan Property \& Casualty Research (HPL).










\printcredits

\bibliographystyle{unsrt}
\bibliography{cas-refs}

\begin{thebibliography}{10}

\bibitem{7286808}
Kim Schouten and Flavius Frasincar.
\newblock Survey on aspect-level sentiment analysis.
\newblock {\em IEEE Transactions on Knowledge and Data Engineering}, 28(3):813--830, 2016.

\bibitem{10.1145/3616855.3635775}
Xusheng Zhao, Hao Peng, Qiong Dai, Xu~Bai, Huailiang Peng, Yanbing Liu, Qinglang Guo, and Philip~S. Yu.
\newblock Rdgcn: Reinforced dependency graph convolutional network for aspect-based sentiment analysis.
\newblock In {\em Proceedings of the 17th ACM International Conference on Web Search and Data Mining}, WSDM '24, page 976–984, New York, NY, USA, 2024. Association for Computing Machinery.

\bibitem{chauhan2023aspect}
Ganpat~Singh Chauhan, Ravi Nahta, Yogesh~Kumar Meena, and Dinesh Gopalani.
\newblock Aspect based sentiment analysis using deep learning approaches: A survey.
\newblock {\em Computer Science Review}, 49:100576, 2023.

\bibitem{li-etal-2019-exploiting}
Xin Li, Lidong Bing, Wenxuan Zhang, and Wai Lam.
\newblock Exploiting {BERT} for end-to-end aspect-based sentiment analysis.
\newblock In {\em Proceedings of the 5th Workshop on Noisy User-generated Text (W-NUT 2019)}, pages 34--41, Hong Kong, China, November 2019. Association for Computational Linguistics.

\bibitem{huang2024flexibly}
Xiaosai Huang, Jing Li, Jia Wu, Jun Chang, Donghua Liu, and Kai Zhu.
\newblock Flexibly utilizing syntactic knowledge in aspect-based sentiment analysis.
\newblock {\em Information Processing \& Management}, 61(3):103630, 2024.

\bibitem{wu2023improving}
Haiyan Wu, Chaogeng Huang, and Shengchun Deng.
\newblock Improving aspect-based sentiment analysis with knowledge-aware dependency graph network.
\newblock {\em Information Fusion}, 92:289--299, 2023.

\bibitem{shi2023syntax}
Jingli Shi, Weihua Li, Quan Bai, Yi~Yang, and Jianhua Jiang.
\newblock Syntax-enhanced aspect-based sentiment analysis with multi-layer attention.
\newblock {\em Neurocomputing}, 557:126730, 2023.

\bibitem{10394369}
Zhijun Lil, Zhenyu Yang, Xiaoyang Li, and Yiwen Li.
\newblock Two-stage aspect sentiment quadruple prediction based on mrc and text generation.
\newblock In {\em 2023 IEEE International Conference on Systems, Man, and Cybernetics (SMC)}, pages 2118--2125, 2023.

\bibitem{zhang-etal-2021-towards-generative}
Wenxuan Zhang, Xin Li, Yang Deng, Lidong Bing, and Wai Lam.
\newblock Towards generative aspect-based sentiment analysis.
\newblock In {\em Proceedings of the 59th Annual Meeting of the Association for Computational Linguistics and the 11th International Joint Conference on Natural Language Processing (Volume 2: Short Papers)}, pages 504--510, Online, August 2021. Association for Computational Linguistics.

\bibitem{zhang-etal-2021-aspect-sentiment}
Wenxuan Zhang, Yang Deng, Xin Li, Yifei Yuan, Lidong Bing, and Wai Lam.
\newblock Aspect sentiment quad prediction as paraphrase generation.
\newblock In {\em Proceedings of the 2021 Conference on Empirical Methods in Natural Language Processing}, pages 9209--9219, Online and Punta Cana, Dominican Republic, November 2021. Association for Computational Linguistics.

\bibitem{mao-etal-2022-seq2path}
Yue Mao, Yi~Shen, Jingchao Yang, Xiaoying Zhu, and Longjun Cai.
\newblock {S}eq2{P}ath: Generating sentiment tuples as paths of a tree.
\newblock In {\em Findings of the Association for Computational Linguistics: ACL 2022}, pages 2215--2225, Dublin, Ireland, May 2022. Association for Computational Linguistics.

\bibitem{Chen_Wang_Liu_Wang_2021}
Shaowei Chen, Yu~Wang, Jie Liu, and Yuelin Wang.
\newblock Bidirectional machine reading comprehension for aspect sentiment triplet extraction.
\newblock {\em Proceedings of the AAAI Conference on Artificial Intelligence}, 35(14):12666--12674, May 2021.

\bibitem{phan-ogunbona-2020-modelling}
Minh~Hieu Phan and Philip~O. Ogunbona.
\newblock Modelling context and syntactical features for aspect-based sentiment analysis.
\newblock In Dan Jurafsky, Joyce Chai, Natalie Schluter, and Joel Tetreault, editors, {\em Proceedings of the 58th Annual Meeting of the Association for Computational Linguistics}, pages 3211--3220, Online, July 2020. Association for Computational Linguistics.

\bibitem{10448322}
Yifan Yang, Yice Zhang, and Ruifeng Xu.
\newblock Enhancing generative aspect-based sentiment analysis with relation-level supervision and prompt.
\newblock In {\em ICASSP 2024 - 2024 IEEE International Conference on Acoustics, Speech and Signal Processing (ICASSP)}, pages 10526--10530, 2024.

\bibitem{yu2023syngen}
Chengze Yu, Taiqiang Wu, Jiayi Li, Xingyu Bai, and Yujiu Yang.
\newblock Syngen: A syntactic plug-and-play module for generative aspect-based sentiment analysis.
\newblock In {\em ICASSP 2023-2023 IEEE International Conference on Acoustics, Speech and Signal Processing (ICASSP)}, pages 1--5. IEEE, 2023.

\bibitem{10.1145/1014052.1014073}
Minqing Hu and Bing Liu.
\newblock Mining and summarizing customer reviews.
\newblock In {\em Proceedings of the Tenth ACM SIGKDD International Conference on Knowledge Discovery and Data Mining}, KDD '04, page 168–177, New York, NY, USA, 2004. Association for Computing Machinery.

\bibitem{li-etal-2021-dual-graph}
Ruifan Li, Hao Chen, Fangxiang Feng, Zhanyu Ma, Xiaojie Wang, and Eduard Hovy.
\newblock Dual graph convolutional networks for aspect-based sentiment analysis.
\newblock In {\em Proceedings of the 59th Annual Meeting of the Association for Computational Linguistics and the 11th International Joint Conference on Natural Language Processing (Volume 1: Long Papers)}, pages 6319--6329, Online, August 2021. Association for Computational Linguistics.

\bibitem{luo-etal-2019-doer}
Huaishao Luo, Tianrui Li, Bing Liu, and Junbo Zhang.
\newblock {DOER}: Dual cross-shared {RNN} for aspect term-polarity co-extraction.
\newblock In {\em Proceedings of the 57th Annual Meeting of the Association for Computational Linguistics}, pages 591--601, Florence, Italy, July 2019. Association for Computational Linguistics.

\bibitem{DBLP:conf/aaai/WanYDLQP20}
Hai Wan, Yufei Yang, Jianfeng Du, Yanan Liu, Kunxun Qi, and Jeff~Z. Pan.
\newblock Target-aspect-sentiment joint detection for aspect-based sentiment analysis.
\newblock In {\em The Thirty-Fourth {AAAI} Conference on Artificial Intelligence, {AAAI} 2020, The Thirty-Second Innovative Applications of Artificial Intelligence Conference, {IAAI} 2020, The Tenth {AAAI} Symposium on Educational Advances in Artificial Intelligence, {EAAI} 2020, New York, NY, USA, February 7-12, 2020}, pages 9122--9129. {AAAI} Press, 2020.

\bibitem{stanovich2000individual}
Keith~E Stanovich and Richard~F West.
\newblock Individual differences in reasoning: Implications for the rationality debate?
\newblock {\em Behavioral and brain sciences}, 23(5):645--665, 2000.

\bibitem{cai-etal-2021-aspect}
Hongjie Cai, Rui Xia, and Jianfei Yu.
\newblock Aspect-category-opinion-sentiment quadruple extraction with implicit aspects and opinions.
\newblock In {\em Proceedings of the 59th Annual Meeting of the Association for Computational Linguistics and the 11th International Joint Conference on Natural Language Processing (Volume 1: Long Papers)}, pages 340--350, Online, August 2021. Association for Computational Linguistics.

\bibitem{hu-etal-2022-improving-aspect}
Mengting Hu, Yike Wu, Hang Gao, Yinhao Bai, and Shiwan Zhao.
\newblock Improving aspect sentiment quad prediction via template-order data augmentation.
\newblock In {\em Proceedings of the 2022 Conference on Empirical Methods in Natural Language Processing}, pages 7889--7900, Abu Dhabi, United Arab Emirates, December 2022. Association for Computational Linguistics.

\bibitem{gou-etal-2023-mvp}
Zhibin Gou, Qingyan Guo, and Yujiu Yang.
\newblock {M}v{P}: Multi-view prompting improves aspect sentiment tuple prediction.
\newblock In {\em Proceedings of the 61st Annual Meeting of the Association for Computational Linguistics (Volume 1: Long Papers)}, pages 4380--4397, Toronto, Canada, July 2023. Association for Computational Linguistics.

\bibitem{zhang-etal-2019-aspect}
Chen Zhang, Qiuchi Li, and Dawei Song.
\newblock Aspect-based sentiment classification with aspect-specific graph convolutional networks.
\newblock In {\em Proceedings of the 2019 Conference on Empirical Methods in Natural Language Processing and the 9th International Joint Conference on Natural Language Processing (EMNLP-IJCNLP)}, pages 4568--4578, Hong Kong, China, November 2019. Association for Computational Linguistics.

\bibitem{tang-etal-2020-dependency}
Hao Tang, Donghong Ji, Chenliang Li, and Qiji Zhou.
\newblock Dependency graph enhanced dual-transformer structure for aspect-based sentiment classification.
\newblock In Dan Jurafsky, Joyce Chai, Natalie Schluter, and Joel Tetreault, editors, {\em Proceedings of the 58th Annual Meeting of the Association for Computational Linguistics}, pages 6578--6588, Online, July 2020. Association for Computational Linguistics.

\bibitem{liang-etal-2022-bisyn}
Shuo Liang, Wei Wei, Xian-Ling Mao, Fei Wang, and Zhiyong He.
\newblock {B}i{S}yn-{GAT}+: Bi-syntax aware graph attention network for aspect-based sentiment analysis.
\newblock In {\em Findings of the Association for Computational Linguistics: ACL 2022}, pages 1835--1848, Dublin, Ireland, May 2022. Association for Computational Linguistics.

\bibitem{gu2023integrating}
Tiquan Gu, Hui Zhao, Zhenzhen He, Min Li, and Di~Ying.
\newblock Integrating external knowledge into aspect-based sentiment analysis using graph neural network.
\newblock {\em Knowledge-Based Systems}, 259:110025, 2023.

\bibitem{10084294}
Aditi Tiwari, Khushboo Tewari, Sukriti Dawar, Ankita Singh, and Nisha Rathee.
\newblock Comparative analysis on aspect-based sentiment using bert.
\newblock In {\em 2023 7th International Conference on Computing Methodologies and Communication (ICCMC)}, pages 723--727, 2023.

\bibitem{yadav2021positionless}
Rohan~Kumar Yadav, Lei Jiao, Morten Goodwin, and Ole-Christoffer Granmo.
\newblock Positionless aspect based sentiment analysis using attention mechanism.
\newblock {\em Knowledge-Based Systems}, 226:107136, 2021.

\bibitem{zhang2022complete}
Hua Zhang, Zeqi Chen, Bi~Chen, Biao Hu, Mian Li, Cheng Yang, and Bo~Jiang.
\newblock Complete quadruple extraction using a two-stage neural model for aspect-based sentiment analysis.
\newblock {\em Neurocomputing}, 492:452--463, 2022.

\bibitem{chen-etal-2022-enhanced}
Hao Chen, Zepeng Zhai, Fangxiang Feng, Ruifan Li, and Xiaojie Wang.
\newblock Enhanced multi-channel graph convolutional network for aspect sentiment triplet extraction.
\newblock In {\em Proceedings of the 60th Annual Meeting of the Association for Computational Linguistics (Volume 1: Long Papers)}, pages 2974--2985, Dublin, Ireland, May 2022. Association for Computational Linguistics.

\bibitem{gao-etal-2022-lego}
Tianhao Gao, Jun Fang, Hanyu Liu, Zhiyuan Liu, Chao Liu, Pengzhang Liu, Yongjun Bao, and Weipeng Yan.
\newblock {LEGO}-{ABSA}: A prompt-based task assemblable unified generative framework for multi-task aspect-based sentiment analysis.
\newblock In {\em Proceedings of the 29th International Conference on Computational Linguistics}, pages 7002--7012, Gyeongju, Republic of Korea, October 2022. International Committee on Computational Linguistics.

\bibitem{lewis2019bart}
Mike Lewis, Yinhan Liu, Naman Goyal, Marjan Ghazvininejad, Abdelrahman Mohamed, Omer Levy, Ves Stoyanov, and Luke Zettlemoyer.
\newblock Bart: Denoising sequence-to-sequence pre-training for natural language generation, translation, and comprehension.
\newblock {\em arXiv preprint arXiv:1910.13461}, 2019.

\bibitem{fei2022lasuie}
Hao Fei, Shengqiong Wu, Jingye Li, Bobo Li, Fei Li, Libo Qin, Meishan Zhang, Min Zhang, and Tat-Seng Chua.
\newblock Lasuie: Unifying information extraction with latent adaptive structure-aware generative language model.
\newblock {\em Advances in Neural Information Processing Systems}, 35:15460--15475, 2022.

\bibitem{10.5555/3455716.3455856}
Colin Raffel, Noam Shazeer, Adam Roberts, Katherine Lee, Sharan Narang, Michael Matena, Yanqi Zhou, Wei Li, and Peter~J. Liu.
\newblock Exploring the limits of transfer learning with a unified text-to-text transformer.
\newblock {\em J. Mach. Learn. Res.}, 21(1), jan 2020.

\bibitem{velivckovic2017graph}
Petar Veli{\v{c}}kovi{\'c}, Guillem Cucurull, Arantxa Casanova, Adriana Romero, Pietro Lio, and Yoshua Bengio.
\newblock Graph attention networks.
\newblock {\em arXiv preprint arXiv:1710.10903}, 2017.

\bibitem{pontiki-etal-2014-semeval}
Maria Pontiki, Dimitris Galanis, John Pavlopoulos, Harris Papageorgiou, Ion Androutsopoulos, and Suresh Manandhar.
\newblock {S}em{E}val-2014 task 4: Aspect based sentiment analysis.
\newblock In {\em Proceedings of the 8th International Workshop on Semantic Evaluation ({S}em{E}val 2014)}, pages 27--35, Dublin, Ireland, August 2014. Association for Computational Linguistics.

\bibitem{pontiki-etal-2015-semeval}
Maria Pontiki, Dimitris Galanis, Haris Papageorgiou, Suresh Manandhar, and Ion Androutsopoulos.
\newblock {S}em{E}val-2015 task 12: Aspect based sentiment analysis.
\newblock In {\em Proceedings of the 9th International Workshop on Semantic Evaluation ({S}em{E}val 2015)}, pages 486--495, Denver, Colorado, June 2015. Association for Computational Linguistics.

\bibitem{pontiki-etal-2016-semeval}
Maria Pontiki, Dimitris Galanis, Haris Papageorgiou, Ion Androutsopoulos, Suresh Manandhar, Mohammad AL-Smadi, Mahmoud Al-Ayyoub, Yanyan Zhao, Bing Qin, Orph{\'e}e De~Clercq, V{\'e}ronique Hoste, Marianna Apidianaki, Xavier Tannier, Natalia Loukachevitch, Evgeniy Kotelnikov, Nuria Bel, Salud~Mar{\'\i}a Jim{\'e}nez-Zafra, and G{\"u}l{\c{s}}en Eryi{\u{g}}it.
\newblock {S}em{E}val-2016 task 5: Aspect based sentiment analysis.
\newblock In {\em Proceedings of the 10th International Workshop on Semantic Evaluation ({S}em{E}val-2016)}, pages 19--30, San Diego, California, June 2016. Association for Computational Linguistics.

\bibitem{xu-etal-2020-position}
Lu~Xu, Hao Li, Wei Lu, and Lidong Bing.
\newblock Position-aware tagging for aspect sentiment triplet extraction.
\newblock In {\em Proceedings of the 2020 Conference on Empirical Methods in Natural Language Processing (EMNLP)}, pages 2339--2349, Online, November 2020. Association for Computational Linguistics.

\bibitem{Li_Bing_Li_Lam_2019}
Xin Li, Lidong Bing, Piji Li, and Wai Lam.
\newblock A unified model for opinion target extraction and target sentiment prediction.
\newblock {\em Proceedings of the AAAI Conference on Artificial Intelligence}, 33(01):6714--6721, Jul. 2019.

\bibitem{xu-etal-2021-learning}
Lu~Xu, Yew~Ken Chia, and Lidong Bing.
\newblock Learning span-level interactions for aspect sentiment triplet extraction.
\newblock In Chengqing Zong, Fei Xia, Wenjie Li, and Roberto Navigli, editors, {\em Proceedings of the 59th Annual Meeting of the Association for Computational Linguistics and the 11th International Joint Conference on Natural Language Processing (Volume 1: Long Papers)}, pages 4755--4766, Online, August 2021. Association for Computational Linguistics.

\bibitem{chen-etal-2022-span}
Yuqi Chen, Chen Keming, Xian Sun, and Zequn Zhang.
\newblock A span-level bidirectional network for aspect sentiment triplet extraction.
\newblock In Yoav Goldberg, Zornitsa Kozareva, and Yue Zhang, editors, {\em Proceedings of the 2022 Conference on Empirical Methods in Natural Language Processing}, pages 4300--4309, Abu Dhabi, United Arab Emirates, December 2022. Association for Computational Linguistics.

\bibitem{chen-qian-2020-relation}
Zhuang Chen and Tieyun Qian.
\newblock Relation-aware collaborative learning for unified aspect-based sentiment analysis.
\newblock In Dan Jurafsky, Joyce Chai, Natalie Schluter, and Joel Tetreault, editors, {\em Proceedings of the 58th Annual Meeting of the Association for Computational Linguistics}, pages 3685--3694, Online, July 2020. Association for Computational Linguistics.

\bibitem{Mao_Shen_Yu_Cai_2021}
Yue Mao, Yi~Shen, Chao Yu, and Longjun Cai.
\newblock A joint training dual-mrc framework for aspect based sentiment analysis.
\newblock {\em Proceedings of the AAAI Conference on Artificial Intelligence}, 35(15):13543--13551, May 2021.

\bibitem{10013820}
Cao~Duy Hoang, Quang~Vinh Dinh, and Ngoc~Hong Tran.
\newblock Aspect-category-opinion-sentiment extraction using generative transformer model.
\newblock In {\em 2022 RIVF International Conference on Computing and Communication Technologies (RIVF)}, pages 1--6, 2022.

\bibitem{xiong2023bart}
Haoliang Xiong, Zehao Yan, Chuhan Wu, Guojun Lu, Shiguan Pang, Yun Xue, and Qianhua Cai.
\newblock Bart-based contrastive and retrospective network for aspect-category-opinion-sentiment quadruple extraction.
\newblock {\em International Journal of Machine Learning and Cybernetics}, 14(9):3243--3255, 2023.

\bibitem{lv-etal-2023-efficient}
Haoran Lv, Junyi Liu, Henan Wang, Yaoming Wang, Jixiang Luo, and Yaxiao Liu.
\newblock Efficient hybrid generation framework for aspect-based sentiment analysis.
\newblock In Andreas Vlachos and Isabelle Augenstein, editors, {\em Proceedings of the 17th Conference of the European Chapter of the Association for Computational Linguistics}, pages 1007--1018, Dubrovnik, Croatia, May 2023. Association for Computational Linguistics.

\end{thebibliography}

\bio{}
\endbio

\endbio

\appendix
\section{Code Environment}
The Code Environment is listed in Table \ref{tab:Code Environment}.
\begin{table}[pos=!ht]
\centering\caption{Code Environment}\label{tab:Code Environment}
\begin{tabular}{l l} \hline
\textbf{Software}&\\
Pytorch & 1.11.0\\
Pytorch\_lightning & 0.8.1\\
Cuda & 11.3 \\
Transformers & 4.30.2\\
Numpy & 1.22.4 \\
Python & 3.8.10\\
Spacy & 3.5.4 \\
\textbf{hardware}&\\
CPU & Intel(R) Xeon(R) 8352V\\
GPU & RTX 4090\\
Memory & 120GB \\
Disk & 500GB\\

\hline
 \end{tabular}
\end{table}

\section{Notation and Definition}
Notations and definitions are shown in Table \ref{tab:Notations}.
\begin{table}[pos=!ht]
\centering\caption{Notations and Definitions}\label{tab:Notations}\scalebox{0.9}{
\begin{tabular}{l l} \hline
Notation & Definition \\
\textbf{Task}&\\
\emph{a} & aspect term\\
\emph{c} & aspect category\\
\emph{o} & opinion term \\
\emph{s} & sentiment polarity\\
 \textbf{Representation}	&	\\
 \emph{H} & $\emph{H} = \left\{h_i\right\}_{n}$ denotes the contextual\\ &  representations from the encoder	\\
 \emph{G} &$\emph{G} = \left\{g_i\right\}_{n}$ denotes the contextual \\&representations optimized by sequence regulator\\
\textbf{Graph}&\\ 
$\mathcal{G}$ &Dependency tree\\
$\mathcal{V}$ & $\mathcal{V} = \left\{v_i\right\}_{n}$ denotes the node set\\
$\mathcal{E}$ & $\mathcal{E} = \left\{e_i\right\}_{n}$ denotes the link set\\
\textbf{Hyper-parameter}&\\ 
$l$ & length in the sequence regulator\\
$d$ & step in the sequence regulator\\\hline
 \end{tabular}}
\end{table}

\section{Hyper-parameter Settings}
The hyper-parameter settings are shown in Table \ref{tab:Hyper}
\begin{table}[pos=!ht]
\centering\caption{Hyper-parameter Settings}\label{tab:Hyper}
\begin{tabular}{l l} \hline
Hyper-parameter & Value\\
\textbf{Sequence Regulator}&\\
Hidden size & 128\\
Attention Head & 8\\
dropout rate & 0.4 \\
Graph Attention Layer & 2\\
Alpha & 0.05\\
Length $l$ & 128\\
Step $d$ & 1\\

 \textbf{Training}	&	\\
 Batch size & 32 \\
 Training Epoch & 40\\
 Evaluation Epoch & 32\\
 Learning rate & 3e-4\\
 Adam epsilon & 1e-8\\
 Seed & 5,15,25 \\
 Model & T5-base\\\hline
 
 \end{tabular}
\end{table}

\newpage
\section{Rule Definition}\label{rule definition}
The rule definition is shown as Table \ref{tab:rule}. 
\begin{table}[pos=!ht]
\centering\caption{Rule definition}\label{tab:rule}\scalebox{1.0}{
\begin{tabular}{l c c c} \hline
& \multicolumn{3}{c}{Rank}\\
  Dependency relation & Rule-1& Rule-2 & Rule-3 \\\hline
 root	&	1	&	1	&	1	\\
 acl 	&	2	&	30	&	5	\\
 acomp 	&	3	&	2	&	2	\\
 advcl 	&	4	&	29	&	3	\\
 advmod 	&	5	&	3	&	4	\\
 agent 	&	6	&	28	&	6	\\
 amod 	&	7	&	27	&	19	\\
 appos 	&	8	&	45	&	20	\\
          attr 	&	9	&	4	&	7	\\
 auxpass 	&	10	&	11	&	30	\\
 case 	&	11	&	46	&	31	\\
 cc 	&	12	&	24	&	12	\\
 ccomp 	&	13	&	25	&	13	\\
 compound 	&	14	&	26	&	14	\\
 conj 	&	15	&	6	&	10	\\
          csubj 	&	16	&	7	&	32	\\
 csubjpass 	&	17	&	31	&	29	\\
 dative 	&	18	&	32	&	44	\\
 dep 	&	19	&	22	&	15	\\
 det 	&	20	&	8	&	16	\\
 dobj 	&	21	&	42	&	17	\\
 expl 	&	22	&	33	&	33	\\
          intj 	&	23	&	23	&	34	\\
 mark 	&	24	&	35	&	28	\\
 meta 	&	25	&	34	&	35	\\
 neg 	&	26	&	10	&	8	\\
 nmod 	&	27	&	21	&	9	\\
 npadvmod 	&	28	&	36	&	36	\\
 nsubj 	&	29	&	5	&	21	\\
          nsubjpass 	&	30	&	12	&	11	\\
 nummod 	&	31	&	40	&	43	\\
 oprd 	&	32	&	37	&	41	\\
 parataxis 	&	33	&	38	&	42	\\
 pcomp 	&	34	&	13	&	22	\\
 pobj 	&	35	&	19	&	26	\\
 poss 	&	36	&	39	&	27	\\
          preconj 	&	37	&	20	&	23	\\
 predet 	&	38	&	41	&	45	\\
 prep 	&	39	&	9	&	18	\\
 prt 	&	40	&	43	&	38	\\
 punct 	&	41	&	14	&	39	\\
 quantmod 	&	42	&	44	&	40	\\
 relcl 	&	43	&	17	&	37	\\
 xcom 	&	44	&	15	&	46	\\
          aux 	&	45	&	16	&	24	\\
 xcomp 	&	46	&	18	&	25	\\
 self 	&	47	&	47	&	47	\\\hline
\end{tabular}}
\end{table}

\end{document}